\def\paperID{4158}
\def\confName{CVPR}
\def\confYear{2025}

\def\paperTitle{SpectroMotion: Dynamic 3D Reconstruction of Specular Scenes}

\def\authorBlock{
Cheng-De Fan$^1$
\quad
Chen-Wei Chang$^1$
\quad
Yi-Ruei Liu$^{1,2}$
\quad
Jie-Ying Lee$^1$
\\
Jiun-Long Huang$^1$
\quad
Yu-Chee Tseng$^1$
\quad
Yu-Lun Liu$^{1}$\vspace{0.75em}
\\
\centerline{$^1$National Yang Ming Chiao Tung University \quad $^2$University of Illinois Urbana-Champaign}
}

\newif\ifreview 
\newif\ifarxiv \newcommand{\arxiv}{\arxivtrue}
\newif\ifcamera 
\newif\ifrebuttal 

\arxiv % \review OR \arxiv OR \cameraready

\pdfoutput=1
\documentclass[10pt,twocolumn,letterpaper]{article}
\ifreview \usepackage[review]{cvpr} \fi
\ifarxiv \usepackage[pagenumbers]{cvpr} \fi
\ifrebuttal \usepackage[rebuttal]{cvpr} \fi
\ifcamera \usepackage{cvpr} \fi

\usepackage{graphicx}	
\usepackage{amsmath}	
\usepackage{amssymb}	
\usepackage{booktabs}
\usepackage{times}
\usepackage{microtype}
\usepackage{epsfig}
\usepackage{caption}
\usepackage{float}
\usepackage{placeins}
\usepackage{color, colortbl}
\usepackage{stfloats}
\usepackage{enumitem}
\usepackage{tabularx}
\usepackage{xstring}
\usepackage{multirow}
\usepackage{xspace}
\usepackage{url}
\usepackage{subcaption}
\usepackage{xcolor}
\usepackage[hang,flushmargin]{footmisc}
\usepackage{bm}

\ifcamera \usepackage[accsupp]{axessibility} \fi

% ...

\ifarxiv  \fi

\newcommand{\R}[1]{{%
    \textbf{%
        \ifstrequal{#1}{1}{\textcolor{red}{R#1}}{%
        \ifstrequal{#1}{2}{\textcolor{blue}{R#1}}{%
        \ifstrequal{#1}{3}{\textcolor{magenta}{R#1}}{%
        \ifstrequal{#1}{4}{\textcolor{teal}{R#1}}{%
                           \textcolor{cyan}{R#1}%
        }}}}%
    }%
}}
\usepackage{xr-hyper}

\makeatletter
\newcommand*{\addFileDependency}[1]{
  \typeout{(#1)}
  \@addtofilelist{#1}
  \IfFileExists{#1}{}{\typeout{No file #1.}}
}

\makeatother
\newcommand*{\myexternaldocument}[1]{
    \externaldocument{#1}
    \addFileDependency{#1.tex}
    \addFileDependency{#1.aux}
}

\definecolor{cvprblue}{rgb}{0.21,0.49,0.74}
\usepackage[pagebackref,breaklinks,colorlinks,allcolors=cvprblue]{hyperref}
\usepackage[capitalize]{cleveref}
\crefname{section}{Sec.}{Secs.}
\crefname{table}{Table}{Tables}
\crefname{figure}{Fig.}{Figs.}

\ifarxiv \crefname{appendix}{App.}{Apps.}
\else \crefname{appendix}{Suppl.}{Suppls.} \fi

\unless\ifarxiv \myexternaldocument{_supplementary} \fi

\begin{document}
%% TITLE
\title{\paperTitle}
\author{\authorBlock}
\maketitle

\begin{abstract}
% Abstract goes here.

We present SpectroMotion, a novel approach that combines 3D Gaussian Splatting (3DGS) with physically-based rendering (PBR) and deformation fields to reconstruct dynamic specular scenes. Previous methods extending 3DGS to model dynamic scenes have struggled to represent specular surfaces accurately. Our method addresses this limitation by introducing a residual correction technique for accurate surface normal computation during deformation, complemented by a deformable environment map that adapts to time-varying lighting conditions. We implement a coarse-to-fine training strategy significantly enhancing scene geometry and specular color prediction. It is the only existing 3DGS method capable of synthesizing photorealistic real-world dynamic specular scenes, outperforming state-of-the-art methods in rendering complex, dynamic, and specular scenes.
Please see our project page at \href{https://cdfan0627.github.io/spectromotion/}{cdfan0627.github.io/spectromotion}.
\end{abstract}
\section{Introduction}
\label{sec:intro}
\begin{figure}[t]
    \centering
    \includegraphics[width=1\linewidth]{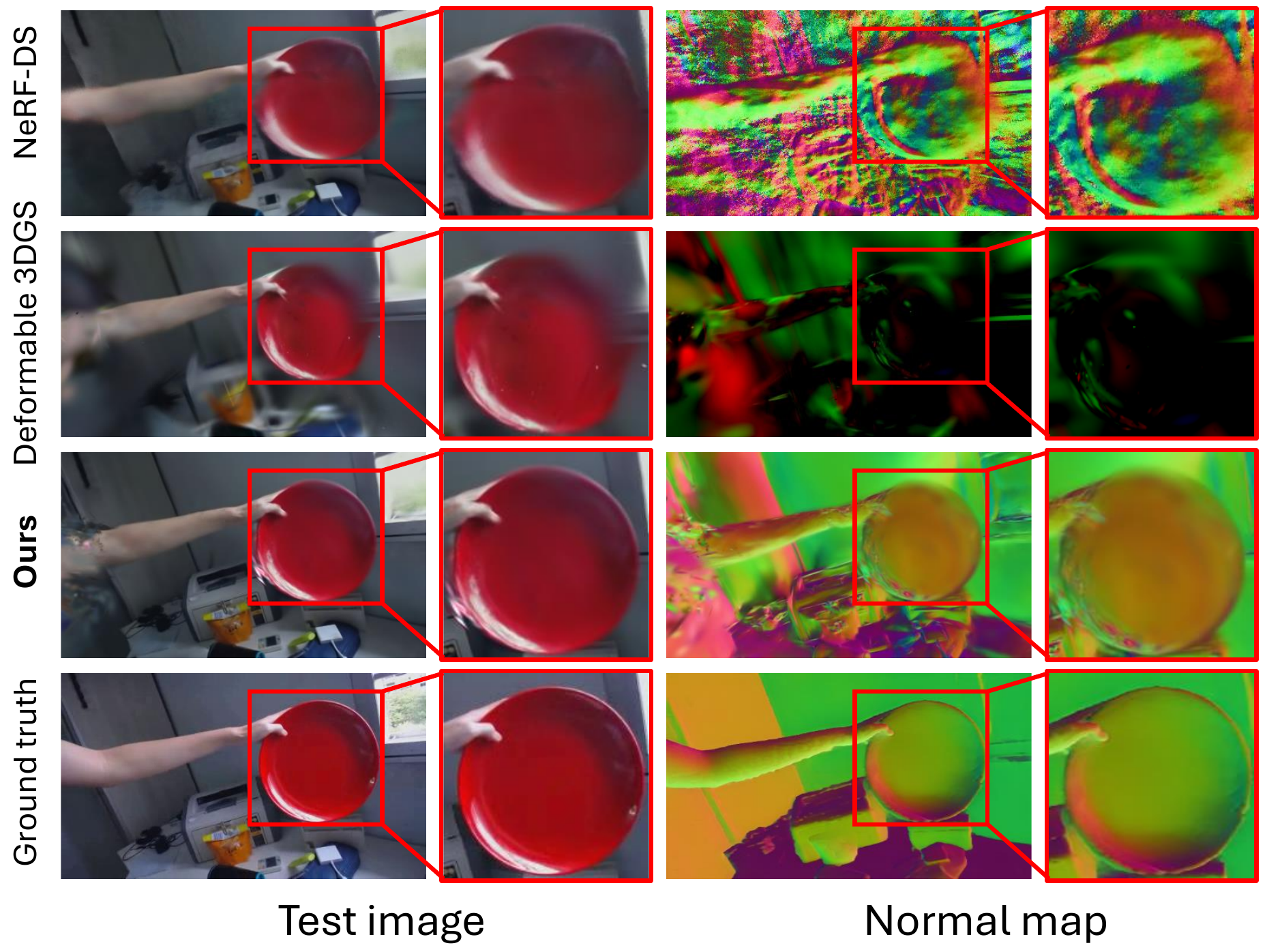}
    \vspace{-4mm}
    \caption{\textbf{Our method, SpectroMotion, recovers and renders dynamic scenes with higher-quality reflections compared to prior work.} It introduces physical normal estimation, deformable environment maps, and a coarse-to-fine training strategy to achieve superior results in rendering dynamic scenes with reflections. Here, we present a rendered test image, corresponding normal maps, and a ground-truth image, where the ground-truth normal map (used as a reference) is generated using a pre-trained normal estimator~\citep{eftekhar2021omnidata}.  For Deformable 3DGS, we use the shortest axes of the deformed 3D Gaussians as the normals. We have highlighted the specular regions to demonstrate the effectiveness of our approach.} 
    \vspace{-2mm}
    \label{fig:teaser}
\end{figure}

3D Gaussian Splatting (3DGS)~\citep{kerbl20233d} has emerged as a groundbreaking technique in 3D scene reconstruction, offering fast training and real-time rendering capabilities. By representing 3D scene using a collection of 3D Gaussians and employing a point-based rendering approach, 3DGS has significantly improved efficiency in novel view synthesis. However, extending 3DGS to model dynamic scenes, especially those containing specular surfaces accurately, has remained a significant challenge.

Existing extensions of 3DGS have made progress in either dynamic scene reconstruction or specular object rendering, but none have successfully combined both aspects. Methods tackling dynamic scenes often struggle with accurately representing specular surfaces, while those focusing on specular rendering are limited to static scenes. This capability gap has hindered the application of 3DGS to real-world scenarios where both motion and specular reflections are present.

We present SpectroMotion, a novel approach that addresses these limitations by combining 3D Gaussian Splatting with physically based rendering (PBR) and deformation fields. Our method introduces three key innovations: a residual correction technique for accurate surface normal computation during deformation, a deformable environment map that adapts to time-varying lighting conditions, and a coarse-to-fine training strategy that significantly enhances scene geometry and specular color prediction.

Our evaluations demonstrate that SpectroMotion outperforms prior methods in view synthesis of scenes containing dynamic specular objects, as illustrated in Fig. \ref{fig:teaser}. It is the only 3DGS method capable of synthesizing photorealistic real-world dynamic specular scenes, surpassing state-of-the-art techniques in rendering complex, dynamic, and specular content. This advancement represents a significant leap in 3D scene reconstruction, particularly for challenging scenarios involving moving specular objects. 

In summary, we make the following contributions:
\begin{itemize}

\item We propose SpectroMotion, a physically-based rendering (PBR) approach combining deformation fields and 3D Gaussian Splatting for real-world dynamic specular scenes.
\item We introduce a residual correction method for accurate surface normals during deformation, coupled with a deformable environment map to handle time-varying lighting conditions in dynamic scenes.
\item We develop a coarse-to-fine training strategy enhancing scene geometry and specular color prediction, outperforming state-of-the-art methods.
\end{itemize}
% To insert a figure: \input{figs/template}
% Or table: \input{tables/template}
\section{Related Work}
\label{sec:related}
\subsection{Dynamic Scene Reconstruction}
Recent works have leveraged NeRF representations to jointly solve for canonical space and deformation fields in dynamic scenes using RGB supervision~\citep{guo2023forward, li2021neural, park2021nerfies, park2021hypernerf, pumarola2020dnerf, tretschk2021nonrigid, xian2021spacetime,liu2023robust,wu2024denver,chen2024narcan,ma2024humannerf}. Further advancements in dynamic neural rendering include object segmentation~\citep{song2023nerfplayer}, incorporation of depth information~\citep{attal2021torf}, utilization of 2D CNNs for scene priors~\citep{lin2022efficient, peng2023representing}, and multi-view video compression~\citep{li2022neural3D}. However, these NeRF-based methods are computationally intensive, limiting their practical applications. To address this, 3D Gaussian Splatting~\citep{kerbl20233d} has emerged as a promising alternative, offering real-time rendering capabilities while maintaining high visual quality. Building upon this efficient representation, recent research has adapted 3D Gaussians for dynamic scenes~\citep{yang2023deformable, wu20234d, huang2024scgs, liang2023gaufre,wang2024shape,mihajlovic2024splatfields,stearns2024dynamic}. Nevertheless, these approaches do not explicitly account for changes in surface normal during the dynamic process. Our work extends this line of research by combining specular object rendering based on normal estimation with a deformation field, enabling each 3D Gaussian to model dynamic specular scenes effectively.

\subsection{Reflective Object Rendering}
While significant progress has been made in rendering reflective objects, challenges from complex light interactions persist. Recent years have seen numerous studies addressing these issues, primarily by decomposing appearance into lighting and material properties ~\citep{bi2020neural, boss2021neuralpil,li2022neural, srinivasan2020nerv, Zhang_2021, Munkberg_2022_CVPR,zhang2021physg,verbin2024eclipse,zhao2024illuminerf}. Building on this foundation, some research has focused on improving the capture and reproduction of specular reflections ~\citep{verbin2022ref, ma2023specnerf, verbin2024nerf, ye20243d}. In contrast, others have leveraged signed distance functions (SDFs) to enhance normal estimation ~\citep{ge2023refneus, liang2023envidr, liang2023spidr, liu2023nero, zhang2023neilf}.
The emergence of 3D Gaussian Splatting (3DGS) has sparked a new wave of techniques ~\citep{jiang2023gaussianshader,liang2023gs,yang2024specgaussian,zhu2024gs,gao2023relightable,shi2023gir} that integrate Gaussian splatting with physically-based rendering. Nevertheless, accurately modeling dynamic environments and time-varying specular reflections remains a significant challenge. To address this limitation, our work introduces a novel approach incorporating a deformable environment map and additional explicit Gaussian attributes specifically designed to capture specular color changes over time.

\section{Method}
\label{sec:method}
\begin{figure*}[t]
    \centering
    \includegraphics[width=1\linewidth]{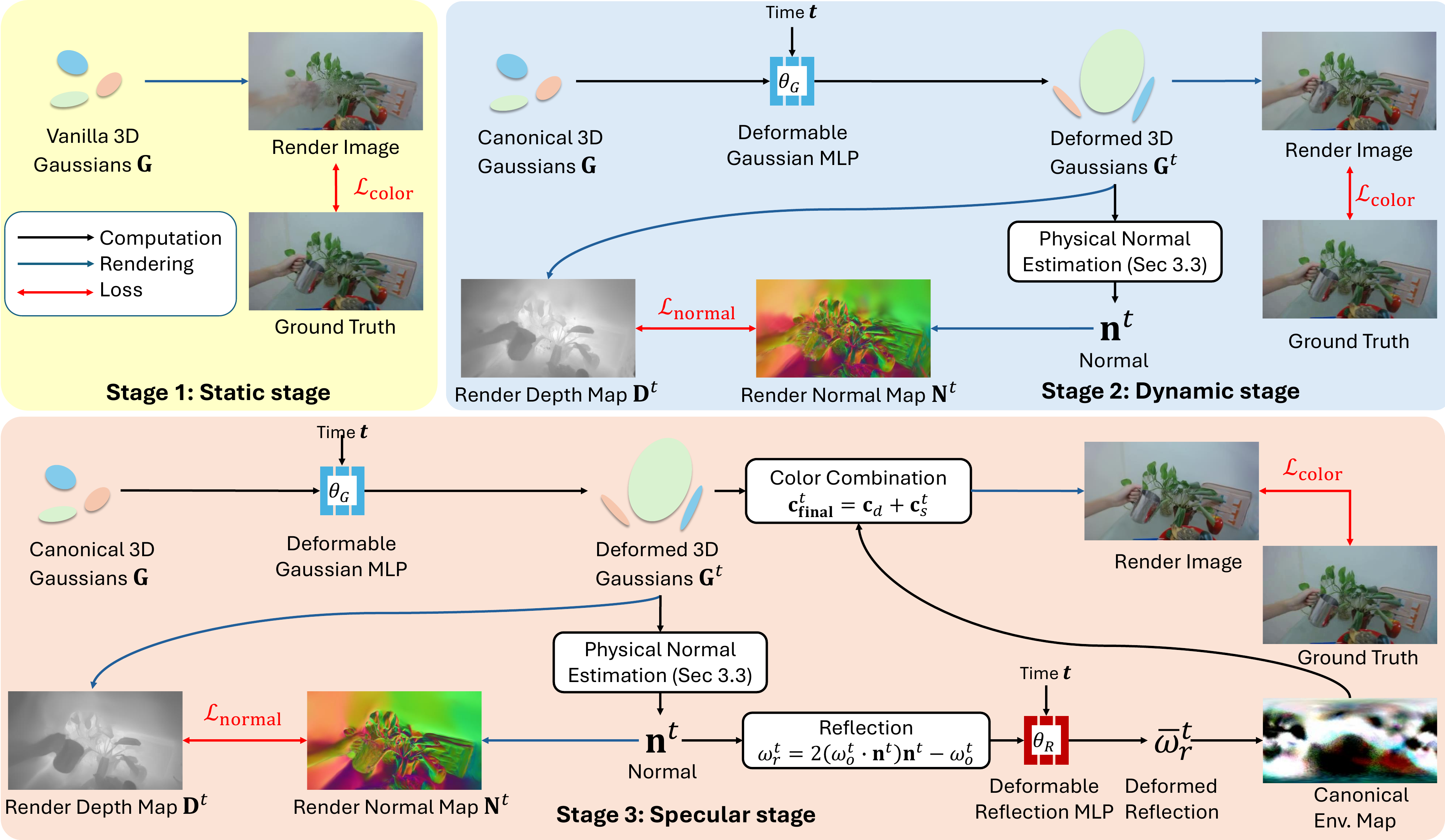}
    \vspace{-4mm}
    \caption{\textbf{Method Overview.} Our method stabilizes the scene geometry through three stages. In the static stage, we stabilize the geometry of the static scene by minimizing photometric loss $\mathcal{L}_{\text{color}}$ between vanilla 3DGS renders and ground truth images. The dynamic stage combines canonical 3D Gaussians $\textbf{G}$ with a deformable Gaussian MLP to model dynamic scenes while simultaneously minimizing normal loss $\mathcal{L}_{\text{normal}}$ between rendered normal map $\mathbf{N}^t$ and gradient normal map from depth map ${\mathbf{D}^t}$, thus further enhancing the overall scene geometry. Finally, the specular stage introduces a deformable reflection MLP to handle changing environment lighting, deforming reflection directions $\omega^t_r$ to query a canonical environment map for specular color $\mathbf{c}_s^t$. It is then combined with diffuse color $\mathbf{c_d}$ (using zero-order spherical harmonics) and learnable specular tint $\mathbf{s_\mathbf{tint}}$ per 3D Gaussian to obtain the final color $\mathbf{c}_\mathbf{final}^t$. This approach enables the modeling of dynamic specular scenes and high-quality novel view rendering.}
    \label{fig:pipeline}
\end{figure*}

\noindent {\bf Overview of the approach.}
The overview of our method is illustrated in Fig.  \ref{fig:pipeline}. Given an input monocular video sequence of frames and corresponding camera poses, we design a three-stage approach to reconstruct the dynamic specular scene, as detailed in Sec.~\ref{sec:Specular Rendering}. Accurate specular reflection requires precise normal estimation, so Sec.~\ref{sec:Normal Estimation} elaborates on how we estimate normals in dynamic scenes. Finally, we introduce the losses used throughout the training process in Sec.~\ref{sec:losses}.

\subsection{Preliminary} \label{sec:Deformation 3D Gaussians}

\vspace{3pt}  \noindent {\bf 3D Gaussian Splatting.}
Each 3D Gaussian is defined by a center position $\bm{x} \in \mathbb{R}^3$ and a covariance matrix $\bm{\Sigma}$. 3D Gaussian Splatting~\citep{kerbl20233d} optimizes the covariance matrix using scaling factors $\bm{s} \in \mathbb{R}^3$ and rotation unit quaternion $\bm{r} \in \mathbb{R}^4$. For novel-view rendering, 3D Gaussians are projected onto 2D camera planes using differentiable splatting~\citep{yifan2019differentiable}:
\begin{equation}
\mathbf{\Sigma}^{\prime}\mathbf{=}\mathbf{J}\mathbf{W}\mathbf{\Sigma}\mathbf{W}^{T}\mathbf{J}^{T}.
\end{equation}
Pixel colors are computed using point-based volumetric rendering:
\begin{equation}
C = \sum_{i \in {N}} {T}_i \mathbf{\alpha}_i {c}_i, \quad \mathbf{\alpha}_i = \mathbf{\sigma}_i e^{-\frac{1}{2} (\bm{x})^T \mathbf{\Sigma}' (\bm{x})},
\end{equation}
where ${T}_i = \prod_{j=1}^{i-1} (1 - \alpha_j)$ is the transmittance, $\sigma_i$ is the opacity, and $\mathbf{c}_i$ is the color of each 3D Gaussian.

\subsection{Specular Rendering}
\label{sec:Specular Rendering}
Since accurate reflections depend heavily on precise geometry, we implement a three-stage coarse-to-fine training strategy: static, dynamic, and specular stages. This approach ensures both stable scene geometry and accurate specular rendering.
% \paragraph{Limitations of existing methods.}
% Current Dynamic 3DGS methods~\citep{wu20234d,yang2023deformable} struggle with specular objects due to low-order spherical harmonics' inability to capture high-frequency details like specular highlights. While some static scene 3DGS-based methods~\citep{jiang2023gaussianshader,liang2023gs} use environment maps to handle specular reflections, these approaches aren't suitable for dynamic scenes. This limitation makes existing 3DGS-based methods inadequate for modeling dynamic scenes with specular objects.

% \paragraph{Proposed solution overview.}
% To overcome these limitations, we propose  physical normal estimation (Sec. \ref{sec:Normal Estimation}) paired with deformable environment maps to model specular color in dynamic scenes. Since accurate reflections depend heavily on precise geometry, we implement a three-stage coarse-to-fine training strategy: static, dynamic, and specular stages. This approach ensures both stable scene geometry and accurate specular rendering.

\subsubsection{Coarse-to-Fine Training Strategy} \label{sec:coarse_to_fine}
\noindent {\bf Static stage.}
In the static stage, we employ vanilla 3DGS~\cite{kerbl20233d} for static scene reconstruction to stabilize the geometry of the static scene. Specifically, we optimize the position $\bm{x}$, scaling $\bm{s}$, rotation $\bm{r}$, opacity $\alpha$, and coefficients of spherical harmonics (SH) of the 3D Gaussians by minimizing the photometric loss $\mathcal{L}_{\text{color}}$ identical to 3DGS~\citep{kerbl20233d}.

\vspace{3pt}  \noindent {\bf Dynamic stage.}
Following the static stage, we address dynamic objects using Deformable 3DGS~\cite{yang2023deformable}. For each 3D Gaussian in canonical 3D Gaussians $\textbf{G}$, we input its position $\bm{x}$ and time $t$ into a deformable Gaussian MLP with parameters $\theta_{\textit{G}}$ to predict position, rotation, and scaling residuals: $(\Delta\bm{x}^t,\Delta\bm{r}^t,\Delta\bm{s}^t) = F_{\theta_G}(\gamma(x), \gamma(t))$, where $\gamma$ denotes positional encoding. 
Attributes of the corresponding 3D Gaussian in deformed 3D Gaussians $\textbf{G}^t$ at time $t$ is obtained by $(\bm{x}^t,\bm{r}^t,\bm{s}^t)=(\Delta\bm{x}^t,\Delta\bm{r}^t,\Delta\bm{s}^t) + (\bm{x},\bm{r},\bm{s}).$

This approach separates motion and geometric structural learning, allowing accurate simulation of dynamic behaviors while maintaining a stable geometric reference.
To further enhance scene geometry, we estimate normals of deformed 3D Gaussians and optimize them using: 
% $\mathcal{L}_{\text{normal}} = 1 - {\mathbf{N}^t}\cdot\hat{\mathbf{N}^t},$ 
\begin{equation}
\mathcal{L}_{\text{normal}} = 1 - {\mathbf{N}^t}\cdot\hat{\mathbf{N}^t},
\end{equation}
where ${\mathbf{N}^t}$ is the rendered normal map and $\hat{\mathbf{N}^t}$ is the normal map derived from the rendered depth map ${\mathbf{D}^t}$. This process improves local associations among 3D Gaussians and optimizes both depth and normal information across the entire scene.

\vspace{3pt}  \noindent {\bf Specular stage.}
We adopt an image-based lighting (IBL) model, where the
environment light is given by a learnable cubemap. Following the rendering equation~\citep{kajiya1986rendering}, split-sum approximation~\citep{karis2013real, Munkberg_2022_CVPR}, and Cook-Torrance reflectance model~\citep{cook1982reflectance}, the outgoing radiance of the specular component $L_s$ is expressed as:
\begin{align}
L_s = \int_{\Omega} & \frac{DGF}{4 (\omega^t_o \cdot \mathbf{n}^t) (\omega_i \cdot \mathbf{n}^t)} 
    (\omega_i \cdot \mathbf{n}^t) d\omega_i \notag \\
    & \times \int_{\Omega} L_i(\omega_i) D(\omega_i, \omega^t_o) 
    (\omega_i \cdot \mathbf{n}^t) d\omega_i,
    \label{eq:specular_reflection}
\end{align}
where $\Omega$ is the hemisphere around the surface normal $\mathbf{n}^t$ (describe in Sec.~\ref{sec:Normal Estimation}.) $D$, $G$, and $F$ represent the GGX normal distribution function~\citep{walter2007microfacet}, geometric attenuation, and Fresnel term, respectively. $\omega^t_o$ is the view direction, and $L_i(\omega_i)$ is the incident radiance.
In the first term, we follow the GaussianShader ~\citep{jiang2024gaussianshader} directly computed by $\mathbf{s}_\text{tint} * F_1 + F_2$, where $F_1$ and $F_2$ are two pre-computed scalars depending on roughness $\rho$, view direction $\omega^t_o$ and normal $\mathbf{n}^t$. Roughness $\rho \in [0, 1]$ and specular tint $\mathbf{s_\mathbf{tint}} \in [0, 1]^3$ are learnable parameters for each 3D Gaussian. The second term is pre-integrated in a filtered learnable cubemap, where each mip-level corresponds to a specific roughness value. The cubemap can be queried using the reflection direction to obtain the value of the second term. After the static and dynamic stages, the geometry is well-defined. This allows us to calculate reflection directions $\omega^t_r = 2 (\omega^t_o \cdot \mathbf{n}^t) \mathbf{n}^t - \omega^t_o$ accurately.

$L_s$ represents only the specular color of the static environment light. To handle time-varying lighting in dynamic scenes, we introduce a deformable environment map, detailed in the following section.

\subsubsection{Deformable Environment Map for Dynamic Lighting.}

The concept of a deformable environment map involves treating the vanilla environment cubemap as a canonical environment map and combining it with a deformation field. This approach enables us to model time-varying lighting conditions effectively. We first apply positional encoding to the reflection direction $\omega^t_r$ and time $t$ to implement this. These encoded values are then input into a deformable reflection MLP with parameters $\theta_{\textit{R}}$. This process allows us to obtain the deformed reflection residual $\Delta\bar\omega^t_r=F_{\theta_{\textit{R}}}(\gamma(\omega^t_r),\gamma(t))$ for each specified time $t$.

Subsequently, we add the deformed reflection residual $\Delta\bar\omega^t_r$ to the reflection direction $\omega^t_r$, yielding the deformed reflection direction $\bar\omega^t_r=\Delta\bar\omega^t_r + \omega^t_r$.

We can then use this deformed reflection direction $\bar\omega^t_r$ to query the canonical environment map. The queried value is then multiplied by the first term of Equation \ref{eq:specular_reflection}, allowing us to obtain time-varying specular color $\mathbf{c}_s^t$. This approach effectively captures the dynamic nature of lighting in the scene while maintaining a stable canonical reference.

\subsubsection{Color Decomposition and Staged Training Strategy.}

We decompose the final color $\mathbf{c}_\mathbf{final}^t$ into diffuse and specular components to better distinguish between high and low-frequency information: $\mathbf{c}_\mathbf{final}^t = \mathbf{c}_d + \mathbf{c}_s^t$,
where $\mathbf{c_d}$ is the diffuse color (using zero-order spherical harmonics as view-independent color), and $\mathbf{c_s}^t$ is the view-dependent color component.
To manage the transition from spherical harmonics to $\mathbf{c}_\mathbf{final}^t$ and mitigate potential geometry disruptions, in the early specular stage, we fix the deformable Gaussian MLP and most 3D Gaussian attributes, optimizing only zero-order SH, specular tint, and roughness. We temporarily suspend densification during this phase. As $\mathbf{c}_\mathbf{final}^t$ becomes more complete, we gradually resume optimization of all parameters and reinstate the densification process.

We further split the specular stage into two parts, applying a coarse-to-fine strategy to the environment map. In the first part, we focus on optimizing the canonical environment map for time-invariant lighting. This establishes a stable foundation for the overall lighting structure. In the second part, we proceed to optimize the deformable reflection MLP for dynamic elements. This approach ensures a more robust learning process, allowing us to capture the static lighting conditions before introducing the complexities of dynamic components.

\begin{figure*}[t]
    \centering
    \includegraphics[width=1\linewidth]{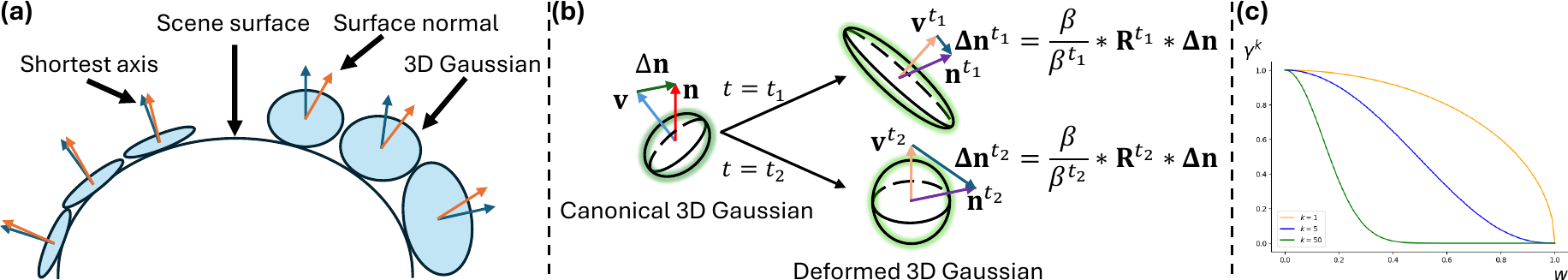}
    \caption{\textbf{Normal estimation.} \textbf{(a)} shows that flatter 3D Gaussians align better with scene surfaces, their shortest axis closely matching the surface normal. In contrast, less flat 3D Gaussians fit less accurately, with their shortest axis diverging from the surface normal. \textbf{(b)} shows that when the deformed 3D Gaussian becomes flatter ($t=t_1$),  normal residual $\Delta\mathbf{n}$ is rotated by $\mathbf{R}^t_1$ and scaled down by $\frac{\beta}{\beta^t_1}$, as flatter Gaussians require smaller normal residuals. Conversely, when the deformation results in a less flat shape ($t=t_2$), $\Delta\mathbf{n}$ is rotated by $\mathbf{R}^t_2$ and amplified by $\frac{\beta}{\beta^t_2}$, requiring a larger correction to align the shortest axis with the surface normal. \textbf{(c)} shows how $\gamma^k$ changes with $w$ (where $w = \frac{|\mathbf{v}_s^t|}{|\mathbf{v}_l^t|}$) for $k=1$, $k=5$, and $k=50$. Larger $w$ indicates less flat Gaussians, while smaller $w$ represents flatter Gaussians. As $k$ increases, $\gamma^k$ decreases more steeply as $w$ rises. For $k=5$, we observe a balanced behavior: $\gamma^k$ approaches 1 for low $w$ and 0 for high $w$, providing a nuanced penalty adjustment across different Gaussian shapes.}
    \label{fig:normal_estimation}
\end{figure*}

\subsection{Physical Normal Estimation}
\label{sec:Normal Estimation}
\noindent {\bf Challenges in normal estimation for 3D Gaussians.}
Normal estimation is essential for modeling specular objects in 3D Gaussians, where GaussianShader \citep{jiang2023gaussianshader} initially used the shortest axis combined with a residual normal for approximation. While this works for static scenes, it becomes problematic with deformed Gaussians because the residual should vary at each time step. A straightforward approach of rotating the residual normal based on quaternion differences between canonical and deformed states proves insufficient, as it does not account for shape changes during deformation. When deformation alters the relative axis lengths, the shortest axis assumption breaks down. This highlights the need for a more comprehensive approach that considers both rotational and shape deformation effects to achieve accurate normal estimation for dynamic specular objects.

\vspace{3pt}  \noindent {\bf Improved rotation calculation for deformed 3D Gaussians.}
To overcome the limitations of naive methods and accurately model the normal of deformed 3D Gaussians, we propose using both the shortest and longest axes of canonical and deformed Gaussians to compute the rotation matrix. This approach accounts for both rotation and shape changes during deformation.
We first align the deformed Gaussian's axes with those of the canonical Gaussian using the following method:
\begin{align}
\mathbf{v}_s^t =
\begin{cases}
\mathbf{v}_s^t & \text{if } \mathbf{v}_s \cdot \mathbf{v}_s^t > 0, \\
-\mathbf{v}_s^t & \text{otherwise.}
\end{cases}
, \quad
\mathbf{v}_l^t =
\begin{cases}
\mathbf{v}_l^t & \text{if } \mathbf{v}_l \cdot \mathbf{v}_l^t > 0, \\
-\mathbf{v}_l^t & \text{otherwise.}
\end{cases},
\end{align}
where $\mathbf{v}_s$ and $\mathbf{v}_l$ represent the shortest and longest axes of canonical 3D Gaussians, while $\mathbf{v}_s^t$ and $\mathbf{v}_l^t$ denote the same for deformed 3D Gaussians. We then construct orthogonal matrices using these aligned axes and their cross-products:
\begin{equation}
\mathbf{U} = \begin{bmatrix} \mathbf{v}_s & \mathbf{v}_l & \mathbf{v}_s \times  \mathbf{v}_l \end{bmatrix},
\quad
\mathbf{V}^t = \begin{bmatrix} \mathbf{v}_s^t & \mathbf{v}_l^t & \mathbf{v}_s^t \times \mathbf{v}_l^t \end{bmatrix}.
\end{equation}
Finally, we derive the rotation matrix $\mathbf{R}^t = \mathbf{V}^t \mathbf{U}^\top.$

% This method provides a robust solution for calculating the rotation of deformation process, ensuring accurate normal estimation for dynamic specular objects.

\vspace{3pt}  \noindent {\bf Adjusting normal residuals and ensuring accuracy.}
To account for shape changes during deformation, we scale the normal residual based on the ratio of oblateness $\frac{\beta}{\beta^t}$ between canonical and deformed 3D Gaussians.
\begin{equation}
\beta = \frac{\lvert\mathbf{v}_l\rvert - \lvert\mathbf{v}_s\rvert}{\lvert\mathbf{v}_l\rvert},
\quad
\beta^t = \frac{\lvert\mathbf{v}_l^t\rvert - \lvert\mathbf{v}_s^t\rvert}{\lvert\mathbf{v}_l^t\rvert},
\end{equation}
where $\beta$ and $\beta^t$ represent the oblateness of canonical and deformed 3D Gaussians, respectively. This is because flatter 3D Gaussians tend to align more closely with the surface, meaning their shortest axis becomes more aligned with the surface normal, as shown in Fig. \ref{fig:normal_estimation} \textbf{(a)}. In such cases, less compensation from the normal residual is needed. Conversely, less flat Gaussians require more compensation, as illustrated in Fig. \ref{fig:normal_estimation} \textbf{(b)}. We then obtain deformed normal residuals:
\begin{equation}
\Delta\mathbf{n}^t = \frac{\beta}{\beta^t}\mathbf{R}^t\Delta\mathbf{n}.
\end{equation}
The final normal $\mathbf{n}^t$ is computed by adding this residual to the shortest axis and ensuring outward orientation:
\begin{equation}
\mathbf{n}^t = \Delta\mathbf{n}^t + \mathbf{v}_s^t,
\quad
\mathbf{n}^t =
\begin{cases}
\mathbf{n}^t &  \text{if } \mathbf{n}^t \cdot \omega^t_o > 0, \\
-\mathbf{n}^t & \text{otherwise.}
\end{cases}
\end{equation}
This approach adjusts for Gaussian flatness and ensures accurate normal estimation.

\subsection{Loss Functions} \label{sec:losses}
\noindent {\bf Normal regularization.}
To allow the normal residual to correct the normal while not excessively influencing the optimization of the shortest axis towards the surface normal, we introduce a penalty term for the normal residual:
\begin{equation}
\mathcal{L}_{\text{reg}} = \gamma^k\| \Delta \mathbf{n} \|_2^2
\quad
\text{where}
\quad
\gamma=\sqrt{1 - \frac{\lvert\mathbf{v}_s^t\rvert^2}{\lvert\mathbf{v}_l^t\rvert^2}}.
\end{equation}
In our experiments, we set $k = 5$. When $k=5$, less flatter 3D Gaussians have $\gamma^k$ approaching 0. Their shortest axis aligns poorly with the surface normal, requiring more normal residual correction and smaller penalties. Conversely, flatter Gaussians have $\gamma^k$ near 1. Their shortest axis aligns better with the surface normal, needing less normal residual correction and allowing larger penalties, as shown in Fig. \ref{fig:normal_estimation} \textbf{(c)}.

\vspace{3pt}  \noindent {\bf Total training loss.}
To refine all parameters in the dynamic and specular stages, we employ the total training loss:
\begin{equation}
\mathcal{L}=  \mathcal{L}_{\text{color}} + \lambda_{\text{normal}}\mathcal{L}_{\text{normal}} + \mathcal{L}_{\text{reg}},
\end{equation}
where $\mathcal{L}_{\text{color}}$ and $\mathcal{L}_{\text{normal}}$ are obtained as described in Section \ref{sec:coarse_to_fine}. 
% where $\mathcal{L}_{\text{reg}}$ is obtained as described in Section \ref{sec:Normal Estimation}. 
In our experiments, we set $\lambda_{\text{normal}} = 0.01$. Due to space constraints, complete implementation details are provided in the supplementary materials.

\section{Experiments}
\label{sec:experinment}
\subsection{Evaluation Results}
We evaluate our method on two real-world datasets: NeRF-DS dataset~\citep{yan2023nerf} and HyperNeRF dataset~\citep{park2021hypernerf}. While GaussianShader~\citep{jiang2023gaussianshader} and GS-IR~\citep{liang2023gs} are originally designed for static scenes and are included here only as reference baselines, we train our method and all baseline approaches for 40,000 iterations to ensure fair comparison. 

\begin{table*}[t]
    \setlength{\tabcolsep}{3pt}
    \centering
    \small
    \caption{\textbf{Quantitative comparison on the NeRF-DS~\citep{yan2023nerf} dataset.} We report the average PSNR, SSIM, and LPIPS (VGG) of several previous models on test images. The \colorbox{red!25}{best}, the \colorbox{orange!25}{second best}, and \colorbox{yellow!25}{third best} results are denoted by red, orange, yellow. }
    \label{tab:whole_scene_tab}
\vspace{-2mm}
    \resizebox{\textwidth}{!}{%
    \begin{tabular}{lccccccccccccccc}
    \toprule
     & \multicolumn{3}{c}{As} & &\multicolumn{3}{c}{Basin}& &\multicolumn{3}{c}{Bell}& &\multicolumn{3}{c}{Cup} \\ \cmidrule{2-4} \cmidrule{6-8} \cmidrule{10-12} \cmidrule{14-16}
    Method & PSNR↑ & SSIM↑& LPIPS↓& &PSNR↑ & SSIM↑& LPIPS↓& &PSNR↑ & SSIM↑& LPIPS↓& &PSNR↑ & SSIM↑& LPIPS↓  \\ 
    \midrule
    Deformable 3DGS~\citep{yang2023deformable} &\colorbox{orange!25}{26.04} &\colorbox{orange!25}{0.8805} &\colorbox{orange!25}{0.1850} & &19.53 &0.7855 &\colorbox{orange!25}{0.1924} & &\colorbox{orange!25}{23.96} &0.7945 &0.2767 & &\colorbox{yellow!25}{24.49} &\colorbox{orange!25}{0.8822} &\colorbox{orange!25}{0.1658} \\ 
    4DGS~\citep{wu20234d} &24.85 &0.8632 &\colorbox{yellow!25}{0.2038} & &19.26 &0.7670 &\colorbox{yellow!25}{0.2196} & &\colorbox{yellow!25}{22.86} &\colorbox{yellow!25}{0.8015} &\colorbox{orange!25}{0.2061} & &23.82 &0.8695 &0.1792 \\
    GaussianShader~\citep{jiang2023gaussianshader} &21.89 &0.7739 &0.3620 & &17.79 &0.6670 &0.4187 & &20.69 &\colorbox{orange!25}{0.8169} &0.3024 & &20.40 &0.7437 &0.3385 \\
    GS-IR~\citep{liang2023gs} &21.58 &0.8033 &0.3033 & &18.06 &0.7248 &0.3135 & &20.66 &0.7829 &\colorbox{yellow!25}{0.2603} & &20.34 &0.8193 &0.2719 \\
    NeRF-DS~\citep{yan2023nerf} &\colorbox{yellow!25}{25.34} &\colorbox{yellow!25}{0.8803} &0.2150 & &\colorbox{orange!25}{20.23} &\colorbox{orange!25}{0.8053} &0.2508 & &22.57 &0.7811 &0.2921 & &\colorbox{orange!25}{24.51} &\colorbox{yellow!25}{0.8802} &\colorbox{yellow!25}{0.1707} \\
    HyperNeRF~\citep{park2021hypernerf} &17.59 &0.8518 &0.2390 & &\colorbox{red!25}{22.58} &\colorbox{red!25}{0.8156} &0.2497 & &19.80 &0.7650 &0.2999 & &15.45 &0.8295 &0.2302 \\
    Ours &\colorbox{red!25}{26.80} &\colorbox{red!25}{0.8843} &\colorbox{red!25}{0.1761} & &\colorbox{yellow!25}{19.75} &\colorbox{yellow!25}{0.7915} &\colorbox{red!25}{0.1896} & &\colorbox{red!25}{25.46} &\colorbox{red!25}{0.8490} &\colorbox{red!25}{0.1600} & &\colorbox{red!25}{24.65} &\colorbox{red!25}{0.8871} &\colorbox{red!25}{0.1588} \\ 
    \midrule
     & \multicolumn{3}{c}{Plate} & &\multicolumn{3}{c}{Press}& &\multicolumn{3}{c}{Sieve}& &\multicolumn{3}{c}{\textbf{Mean}} \\ \cmidrule{2-4} \cmidrule{6-8} \cmidrule{10-12} \cmidrule{14-16}
    Method & PSNR↑ & SSIM↑& LPIPS↓& &PSNR↑ & SSIM↑& LPIPS↓& &PSNR↑ & SSIM↑& LPIPS↓& &PSNR↑ & SSIM↑& LPIPS↓  \\ 
    \midrule
    Deformable 3DGS~\citep{yang2023deformable} &19.07 &0.7352 &0.3599 & &\colorbox{orange!25}{25.52} &\colorbox{yellow!25}{0.8594} &\colorbox{orange!25}{0.1964} & &\colorbox{red!25}{25.37} &\colorbox{orange!25}{0.8616} &\colorbox{orange!25}{0.1643} & &\colorbox{orange!25}{23.43} &\colorbox{yellow!25}{0.8284} &\colorbox{yellow!25}{0.2201} \\ 
    4DGS~\citep{wu20234d} &18.77 &0.7709 &\colorbox{orange!25}{0.2721} & &24.82 &0.8355 &\colorbox{yellow!25}{0.2255} & &\colorbox{yellow!25}{25.16} &0.8566 &\colorbox{yellow!25}{0.1745} & &22.79 &0.8235 &\colorbox{orange!25}{0.2115} \\
    GaussianShader~\citep{jiang2023gaussianshader} &14.55 &0.6423 &0.4955 & &19.97 &0.7244 &0.4507 & &22.58 &0.7862 &0.3057 & &19.70 &0.7363 &0.3819 \\
    GS-IR~\citep{liang2023gs} &15.98 &0.6969 &0.4200 & &22.28 &0.8088 &0.3067 & &22.84 &0.8212 &0.2236 & &20.25 &0.7796 &0.2999 \\
    NeRF-DS~\citep{yan2023nerf} &\colorbox{yellow!25}{19.70} &\colorbox{yellow!25}{0.7813} &\colorbox{yellow!25}{0.2974} & &\colorbox{yellow!25}{25.35} &\colorbox{red!25}{0.8703} &0.2552 & &24.99 &\colorbox{red!25}{0.8705} &0.2001 & &\colorbox{yellow!25}{23.24} &\colorbox{orange!25}{0.8384 }&0.2402 \\
    HyperNeRF~\citep{park2021hypernerf} &\colorbox{red!25}{21.22} &\colorbox{orange!25}{0.7829} &0.3166 & &16.54 &0.8200 &0.2810 & &19.92 &0.8521 &0.2142 & &19.01 &0.8167 &0.2615 \\
    Ours &\colorbox{orange!25}{20.84} &\colorbox{red!25}{0.8172} &\colorbox{red!25}{0.2198} & &\colorbox{red!25}{26.49} &\colorbox{orange!25}{0.8657} &\colorbox{red!25}{0.1889} & &\colorbox{orange!25}{25.22} &\colorbox{red!25}{0.8705} &\colorbox{red!25}{0.1513} & &\colorbox{red!25}{24.17} &\colorbox{red!25}{0.8522 }&\colorbox{red!25}{0.1778} \\ 
    \bottomrule
    \end{tabular}
    }
\end{table*}
\begin{figure*}[t]
    \centering
    \includegraphics[width=1\linewidth]{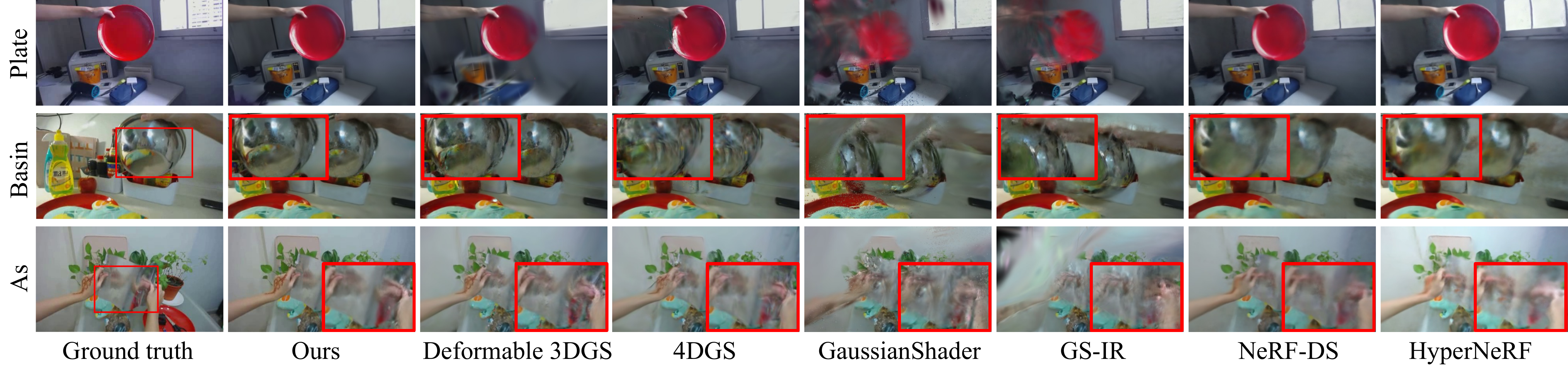}
    \vspace{-4mm}
    \caption{\textbf{Qualitative comparison on the NeRF-DS~\cite{yan2023nerf} dataset.}}
    \label{fig:whole_scene}
\end{figure*} 

\vspace{3pt}  \noindent {\bf NeRF-DS dataset.}
 The NeRF-DS dataset \citep{yan2023nerf} is a monocular video dataset comprising seven real-world scenes from daily life featuring various types of moving or deforming specular objects. We compare our method with the most relevant state-of-the-art approaches. As shown in Tab. \ref{tab:whole_scene_tab} and Fig. \ref{fig:whole_scene}, the quantitative results demonstrate that our method decisively outperforms baselines in reconstructing and rendering real-world highly reflective dynamic specular scenes. 
 
 The rendering speed is correlated with the quantity of 3D Gaussians. When the number of 3D Gaussians is below 178k, our method can achieve real-time rendering over 30 FPS on an NVIDIA RTX 4090.

\begin{table}[t]
\centering
\small
\caption{\textbf{Quantitative comparison on the HyperNeRF~\citep{park2021hypernerf} dataset.} We report the average PSNR, SSIM, and LPIPS (VGG) of several previous models. The \colorbox{red!25}{best}, the \colorbox{orange!25}{second best}, and \colorbox{yellow!25}{third best} results are denoted by red, orange, yellow.}
\label{tab:hyper_main}
\vspace{-2mm}
\begin{tabular}{lccc}
\toprule
Method & PSNR ↑ & SSIM ↑ & LPIPS ↓ \\
\midrule
Deformable 3DGS~\citep{yang2023deformable} &{22.78} &{0.6201} &\colorbox{orange!25}{0.3380} \\
4DGS~\citep{wu20234d} &\colorbox{red!25}{24.89} &\colorbox{red!25}{0.6781} &\colorbox{yellow!25}{0.3422}  \\
GaussianShader~\citep{jiang2023gaussianshader} &18.55 &0.5452 &0.4795  \\
GS-IR~\citep{liang2023gs} &19.87 &0.5729 &0.4498 \\
NeRF-DS~\citep{yan2023nerf} &\colorbox{orange!25}{23.65} &\colorbox{orange!25}{0.6405} &0.3972\\
HyperNeRF~\citep{park2021hypernerf}
&\colorbox{yellow!25}{23.11} &\colorbox{yellow!25}{0.6387} &{0.3968}\\
Ours &22.22 &0.6088 &\colorbox{red!25}{0.3295}\\
\bottomrule
\end{tabular}
\end{table}
\begin{figure*}[t]
    \centering
    \includegraphics[width=1\linewidth]{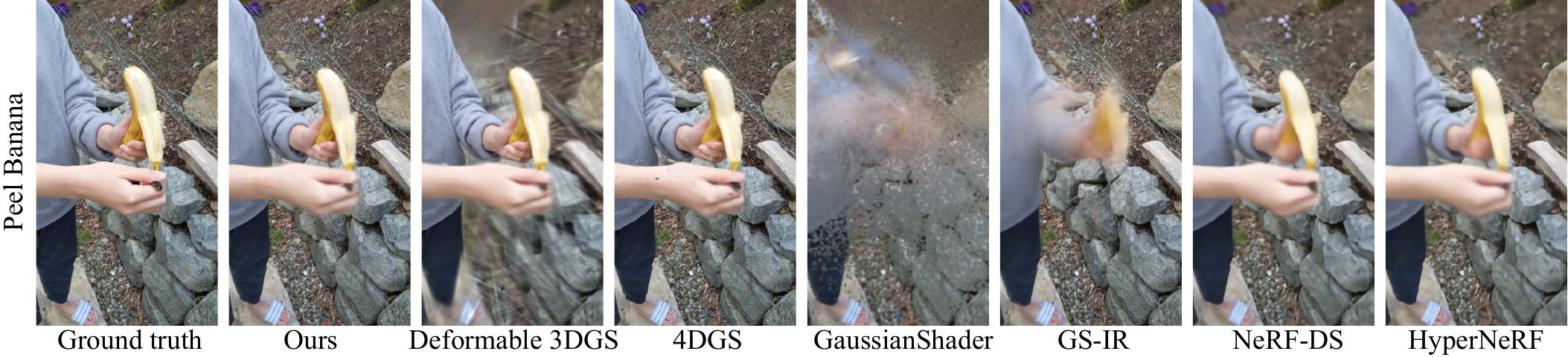}
    \vspace{-4mm}
    \caption{\textbf{Qualitative comparison on the HyperNeRF~\cite{park2021hypernerf} dataset.}}
    \label{fig:hypernerf}
\end{figure*} 

\vspace{3pt}  \noindent {\bf HyperNeRF dataset.}
The HyperNeRF dataset contains real-world dynamic scenes and does not include specular objects. As shown in Tab. \ref{tab:hyper_main} and Fig. \ref{fig:hypernerf}, the results demonstrate that we are on par with state-of-the-art techniques for rendering novel views, and our method's performance is not limited to shiny scenes.

\begin{figure}[t]
    \includegraphics[width=1\linewidth]{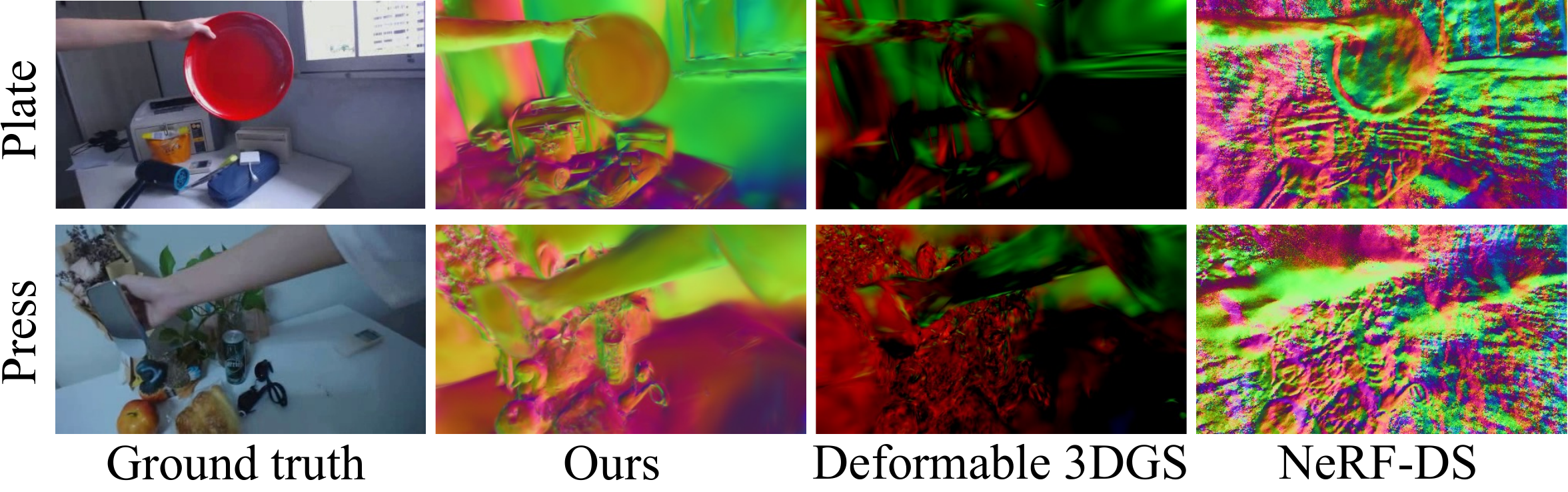}
    \vspace{-4mm}
    \caption{\textbf{Qualitative comparison of normal maps between our method, Deformable 3DGS, and NeRF-DS.}}
    \label{fig:normal_compared}
\end{figure} 
% Notably, unlike NeRF-DS, our approach does not require mask supervision to clearly distinguish between static and dynamic objects, as illustrated in Fig.~\ref{fig:dynamic_mask}. Additionally, Fig.~\ref{fig:decompose} illustrates our method's decomposition results. Our approach consistently achieves a realistic separation of specular and diffuse components across different scenes in the NeRF-DS dataset. 
In Fig.~\ref{fig:normal_compared}, we compare our method's normal maps with those from Deformable 3DGS~\citep{yang2023deformable} and NeRF-DS~\citep{yan2023nerf}. For Deformable 3DGS~\citep{yang2023deformable}, we obtain the normals by using the shortest axes of the deformed 3D Gaussians. As demonstrated, our method produces significantly better quality normal maps compared to Deformable 3DGS~\citep{yang2023deformable} and NeRF-DS~\citep{yan2023nerf}.

\subsection{Ablation Study}
For a fair comparison, we train our method and all ablation
experiments for 40,000 iterations.

\begin{table}[t]
\centering
\small
\caption{\textbf{Ablation studies on different coarse to fine training strategy stages.}}
\label{tab:stage}
\vspace{-2mm}
\begin{tabular}{lccc}
\toprule
Stage & PSNR ↑ & SSIM ↑ & LPIPS ↓ \\
\midrule
Static &20.26  & 0.7785 & 0.2953 \\
St. + Dynamic &24.02  &0.8508  &0.1831  \\
St. + Dy. + Specular &\textbf{24.17}  &\textbf{0.8522}  &\textbf{0.1778}  \\
\bottomrule
\end{tabular}
\end{table}

\begin{figure}[t]
    \centering
    \includegraphics[width=1\linewidth]{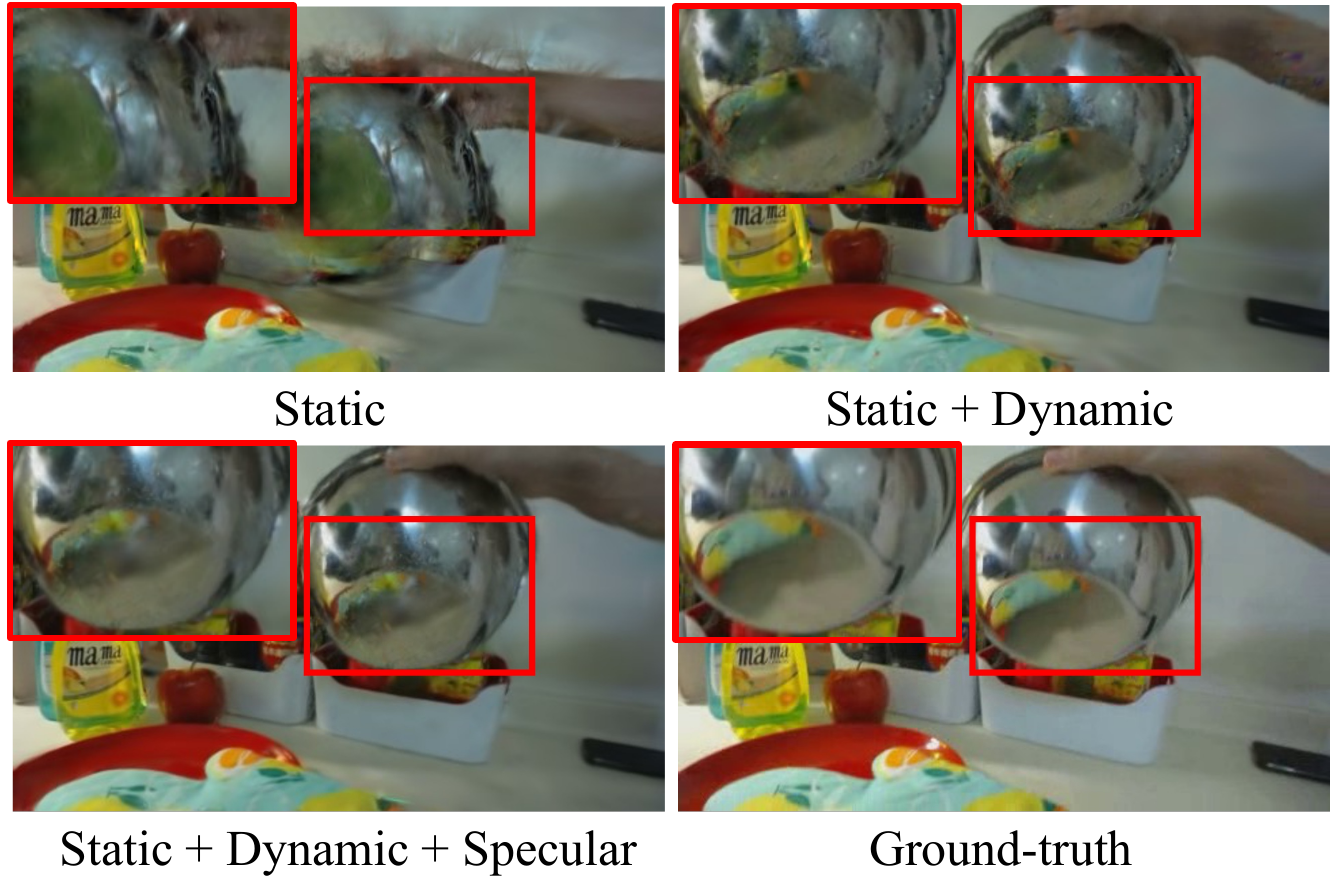}
    \vspace{-4mm}
    \caption{\textbf{Qualitative comparison of each training stage in our coarse-to-fine approach.} 
}
    \label{fig:stage}
\end{figure}

\vspace{3pt}  \noindent {\bf Different coarse to fine training strategy stages.}
As shown in Tab. \ref{tab:stage} and Fig. \ref{fig:stage}, each stage contributes effectively to the model's performance. The Dynamic stage enhances dynamic object stability compared to the Static stage alone, while the Specular stage improves reflection clarity beyond the combined Static and Dynamic stages. 

\begin{table}[t]
\centering
\small
\caption{\textbf{Ablation studies on different coarse to fine training strategy stages.}}
\label{tab:abalation_with_losses}
\vspace{-2mm}
\begin{tabular}{cccc|ccc}
    \toprule
    C2F &  $\mathcal{L}_{\text {normal}}$ & $\mathcal{L}_{\text {reg}}$& $\gamma^k$& PSNR↑ & SSIM↑ & LPIPS↓ \\
    \midrule
    &  \checkmark & \checkmark &\checkmark &23.16&0.8294  &0.2156  \\
    \checkmark &  &  &&23.40  &0.8277  &0.2278  \\
    \checkmark & \checkmark &  &&24.15  &0.8510  &0.1845  \\
    \checkmark & \checkmark & \checkmark &&24.09  &0.8515  &0.1818  \\
    \checkmark & \checkmark & \checkmark &\checkmark&\textbf{24.17}  &\textbf{0.8522}  &\textbf{0.1778}  \\
    \bottomrule
\end{tabular}
\end{table}

\begin{figure*}[t]
    \centering
    \includegraphics[width=1\linewidth]{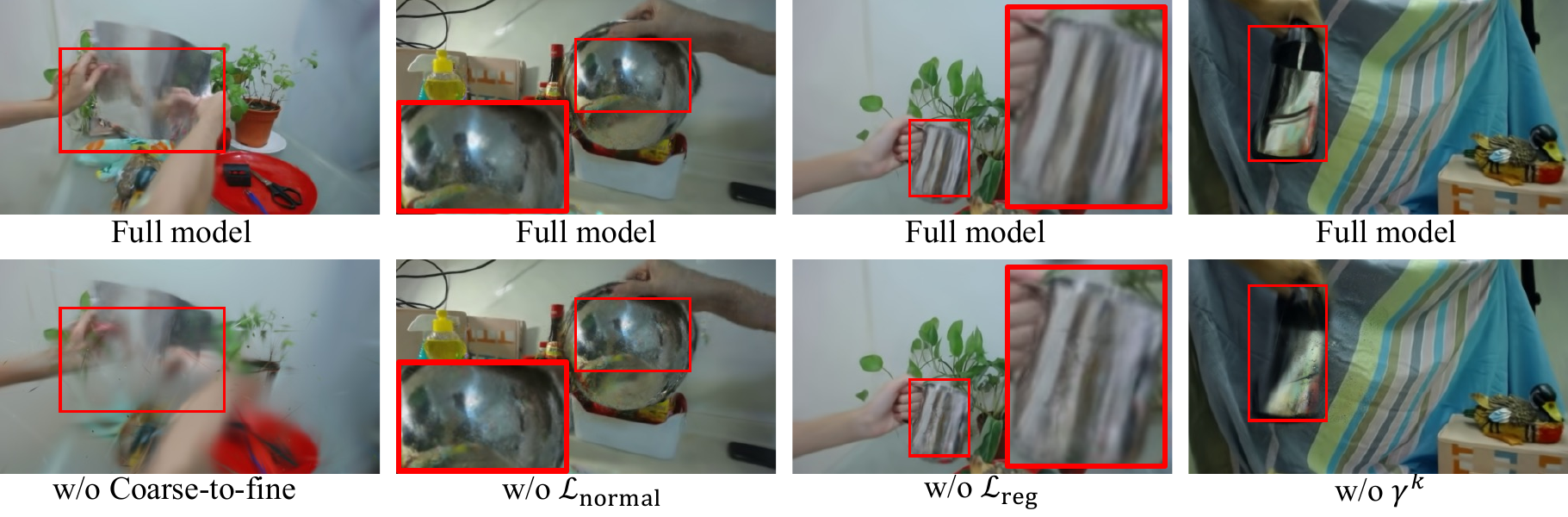}
    \vspace{-4mm}
    \caption{\textbf{Qualitative comparison of ablation study without different components.}}
    \label{fig:ab_study_vis}
\end{figure*} 

\vspace{3pt}  \noindent {\bf Ablation study on coarse-to-fine, and loss function.}
The model's performance was evaluated without key components: the coarse-to-fine training strategy, normal loss $\mathcal{L}_{\text{normal}}$, normal regularization $\mathcal{L}_{\text{reg}}$, and $\gamma^k$. Fig.~\ref{fig:ab_study_vis} and Tab. \ref{tab:abalation_with_losses} illustrate the effects of these omissions.
Without the coarse-to-fine approach, the model, trained directly at the specular stage, produces incomplete scene geometry, affecting environment map queries for specular color. Omitting normal loss $\mathcal{L}_{\text{normal}}$ removes direct supervision on normals and leads to inaccurate reflection directions and less precise specular colors. Removing normal regularization $\mathcal{L}_{\text{reg}}$ allows the normal residual to dominate normal optimization, resulting in insufficient optimization of the 3D Gaussians' shortest axis towards the correct normal, which in turn reduces the rendering quality. The normal residual decreases for non-flattened and flat Gaussians without $\gamma^k$ in normal regularization. This particularly affects less flat  3D Gaussians whose shortest axis significantly deviates from the surface normal. The insufficient normal residual correction causes these 3D Gaussians' shortest axes to deviate greatly from their original direction in an attempt to align with the surface normal, ultimately reducing rendering quality. 

\begin{table}[t]
\centering
\small
\caption{\textbf{Ablation studies on SH, Static and Deformable environment map.}}
\label{tab:abalation_sh}
\vspace{-2mm}
\begin{tabular}{lccc}
    \toprule
      & PSNR ↑ & SSIM ↑ & LPIPS ↓ \\
    \midrule
    SH &23.63  &0.8453  &0.1844  \\
    Static Env. map &24.02  &0.8508  &0.1831  \\
    Deformable Env. map &\textbf{24.17}  &\textbf{0.8522}  &\textbf{0.1778}  \\
    \bottomrule
\end{tabular}
\end{table}

\begin{figure}[t]
    \centering
    \includegraphics[width=1\linewidth]{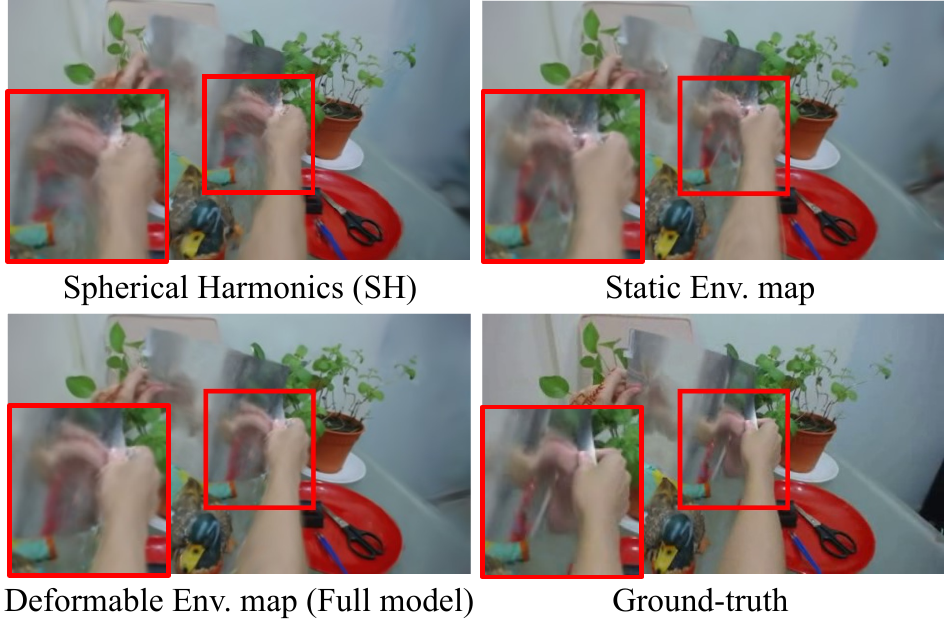}
    \vspace{-4mm}
    \caption{\textbf{Qualitative comparison of ablation study on SH, Static environment map, and Deformable environment map.}}
    \label{fig:env}
\end{figure} 

\vspace{3pt}  \noindent {\bf Ablation study on SH, Static environment map, and Deformable environment map.}
Fig.~\ref{fig:env} and Tab. \ref{tab:abalation_sh} demonstrate the superiority of the deformable environment map over the static environment map, which in turn outperforms Spherical Harmonics (SH). SH struggles to accurately model high-frequency specular colors. While the static environment map can model high-frequency colors, it is best suited for static lighting conditions. In contrast, the deformable environment map models time-varying lighting, offering superior performance for dynamic scenes.

\begin{table}[t]
\centering
\small
\caption{\textbf{Ablation studies on 2DGS and without Physical Normal Estimation.}}
\label{tab:ablation_2dgs}
\vspace{-2mm}
\begin{tabular}{lccc}
    \toprule
      & PSNR ↑ & SSIM ↑ & LPIPS ↓ \\
    \midrule
    2DGS~\citep{huang20242d} &23.22  &0.8219  &0.2283  \\
    w/o N.E. &23.89  &0.8490  &0.1837  \\
    Full model &\textbf{24.17}  &\textbf{0.8522}  &\textbf{0.1778}  \\
    \bottomrule
\end{tabular}
\end{table}
\begin{figure}[t]
    \centering
    \includegraphics[width=1\linewidth]{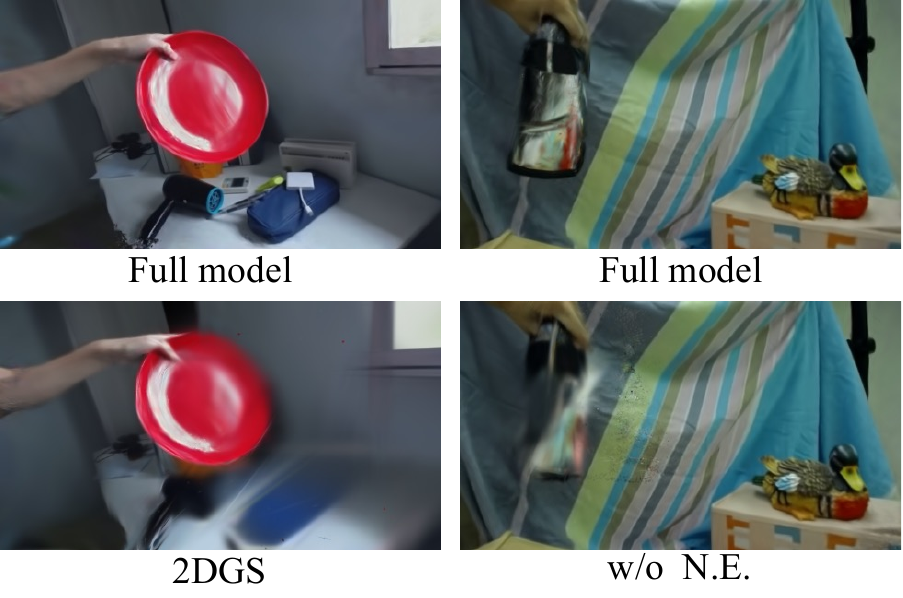}
    \vspace{-4mm}
    \caption{\textbf{Qualitative comparison of ablation study on 2DGS and without Physical Normal Estimation.}}
    \label{fig:ablation_2DGS}
\end{figure} 

\vspace{3pt}  \noindent {\bf Ablation study on 2DGS~\citep{huang20242d} and without Physical Normal Estimation.}
In Fig. \ref{fig:ablation_2DGS} and Tab. \ref{tab:ablation_2dgs}, "2DGS" represents replacing our 3D Gaussian with 2D Gaussian representation. Since 2DGS inherently includes normals, we omit physical normal estimation. "w/o N.E." means skipping physical normal estimation and using the shortest axis of 3D Gaussians as the normal. This causes the normal loss $\mathcal{L}_{\text{normal}}$ to directly supervise the shortest axes, making some axes deviate significantly to align with surface normals, resulting in degraded rendering quality.
\section{Conclusion}
\label{sec:conclusion}
SpectroMotion enhances 3D Gaussian Splatting for dynamic specular scenes by combining specular rendering with deformation fields. Using normal residual correction, coarse-to-fine training, and a deformable environment map, it achieves superior accuracy and visual quality in novel view synthesis, outperforming existing methods while maintaining geometric consistency.

\vspace{3pt}  \noindent {\bf Limitations.}
Though we stabilize the entire scene's geometry using a coarse-to-fine training strategy, some failure cases still occur. Please refer to the supplementary materials for visual results of failure cases.

\paragraph{Acknowledgements.}
This research was funded by the National Science and Technology Council, Taiwan, under Grants NSTC 112-2222-E-A49-004-MY2 and 113-2628-E-A49-023-. The authors are grateful to Google, NVIDIA, and MediaTek Inc. for their generous donations. Yu-Lun Liu acknowledges the Yushan Young Fellow Program by the MOE in Taiwan.

{\small
\bibliographystyle{ieeenat_fullname}
\bibliography{11_references}
}

\ifarxiv \clearpage \appendix \section{Overview}
\label{sec:appendix_section}
Supplementary material goes here.
This supplementary material presents additional results to complement the main manuscript. In Section \ref{sec:Implementation}, we detail our coarse-to-fine training strategy, along with the network architectures of the deformable Gaussian MLP and deformable reflection MLP. Subsequently, Section \ref{sec:result} presents additional results, including comprehensive comparisons, more visualizations, and training and rendering efficiency. Finally, in Section \ref{sec:limit}, we discuss the limitation of our approach and provide visual example of failure case.

\section{Implementation Details}~\label{sec:Implementation} We use PyTorch as our framework and 3DGS~\citep{kerbl20233d} as our codebase. Our coarse-to-fine training strategy is divided into three sequential stages: static, dynamic, and specular stages. 
\paragraph{Static stage.} 
In the static stage, we train the vanilla 3D Gaussian Splatting (3DGS) for 3000 iterations to stabilize the static geometry. 
\paragraph{Dynamic stage.} 
After the static stage, we move on to the dynamic stage. During this phase, we introduce a deformable Gaussian MLP to model dynamic objects. First, we optimize both the canonical Gaussians and the deformable Gaussian MLP for 3,000 iterations until the scene reaches a relatively stable state. Then, we introduce the normal loss $\mathcal{L}_{\text{normal}}$, enabling simultaneous optimization of the scene's normal and depth, and perform an additional 3,000 iterations to further refine the geometry.  The dynamic stage comprises a total of 6,000 training iterations.
\paragraph{Specular stage.} 
After the dynamic stage concludes, we transition to the specular stage, which involves changing the color representation from complete spherical harmonics to $\mathbf{c}_\mathbf{final}$. To mitigate potential geometry disruptions due to the initially incomplete $\mathbf{c}_\mathbf{final}$, we fix the deformable Gaussian MLP and all 3D Gaussian attributes except for zero-order SH, specular tint, and roughness, while temporarily suspending densification. After 6000 iterations, once $\mathbf{c}_\mathbf{final}$ becomes more complete, we resume optimization of all parameters and reinstate the densification process. Then, after another 3000 iterations, we stop the densification process. Concurrently, during the first 2000 iterations of the specular stage, we optimize only the canonical environment map to learn time-invariant lighting. For the canonical environment map, we use $6 \times 128 \times 128$ learnable parameters. Subsequently, we begin optimizing the deformable reflection MLP to capture time-varying lighting effects until the training is complete. The specular stage comprises a total of 31,000 training iterations. For the Peel Banana scene in the HyperNeRF dataset, we do not fix the deformable Gaussian MLP. We resume optimization of all parameters and reinstate the densification process after the first 4000 iterations of the specular stage. Then, after another 2000 iterations, we stop the densification process to prevent excessive growth in the number of 3D Gaussians, which could lead to GPU out-of-memory issues.

For the entire experiment, we train for a total of 40,000 iterations and we use Adam optimizer.

\subsection{Network Architecture of the Deformable Gaussian MLP and Deformable reflection MLP}
\begin{figure}[t]
    \centering
    \includegraphics[width=1\linewidth]{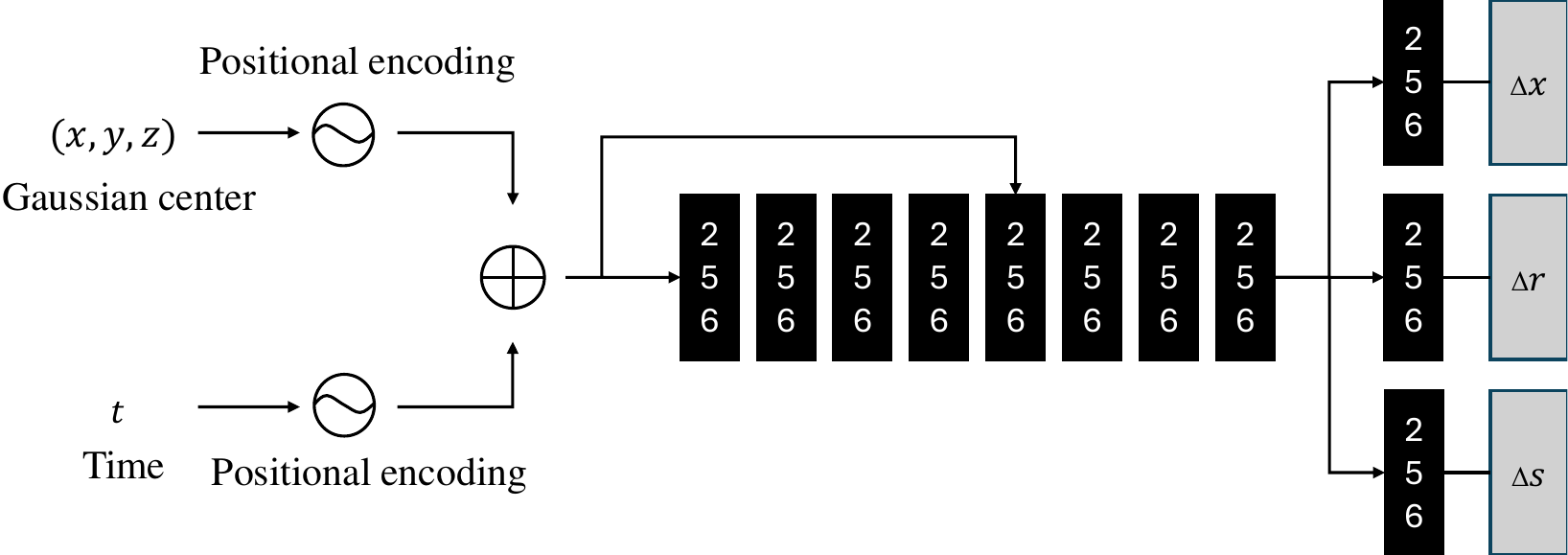}
    \caption{\textbf{ Architecture of the deformable Gaussian MLP}}
    \label{fig:deform_network}
\end{figure}
\begin{figure}[t]
    \centering
    \includegraphics[width=1\linewidth]
    {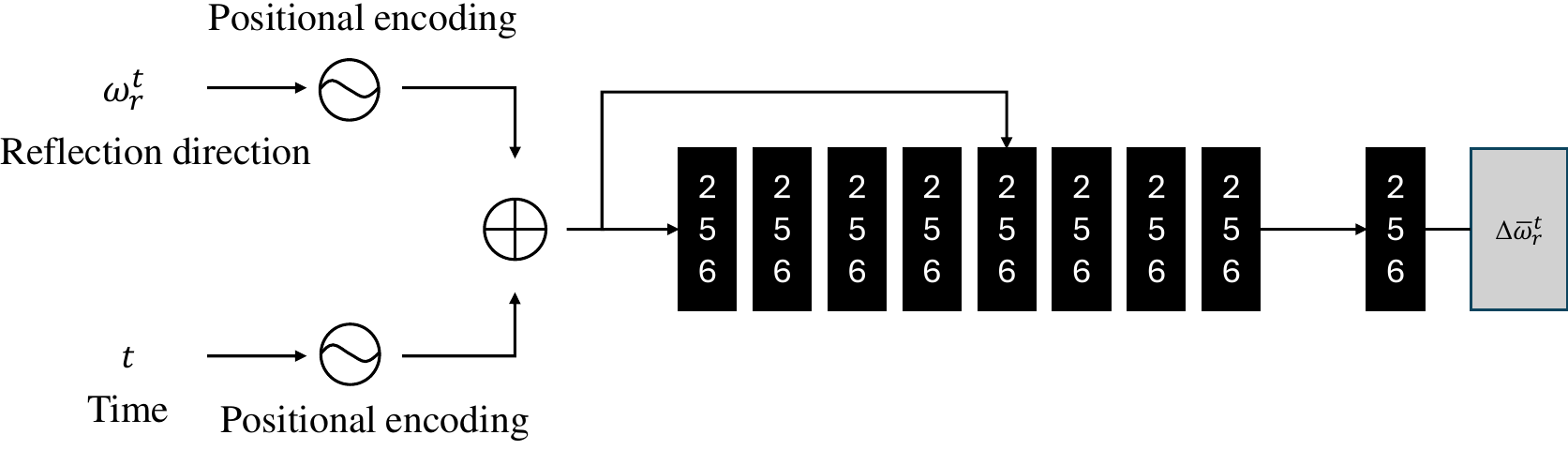}
    \caption{\textbf{ Architecture of the deformable reflection MLP}}
    \label{fig:deform_refl}
\end{figure}
We follow Deformable 3DGS~\citep{yang2023deformable} and use deformable Gaussian MLP to predict each coordinate of 3D Gaussians and time to their corresponding deviations in position, rotation, and scaling. As shown in Fig. ~\ref{fig:deform_network}, the MLP initially processes the input through eight fully connected layers that employ ReLU activations, featuring 256-dimensional hidden layers and outputs a 256-dimensional feature vector. This vector is then passed through three additional fully connected layers combined with ReLU activation to separately output the offsets over time for position, rotation, and scaling. Notably, similar to NeRF, the feature vector and the input are concatenated in the fourth layer. For the deformable reflection MLP, we utilize the same network architecture, as shown in Fig.~\ref{fig:deform_refl}.

\section{Additional Results}~\label{sec:result}
\subsection{Dynamic specular object of NeRF-DS dataset.}
\begin{table*}[t]
    \setlength{\tabcolsep}{3pt}
    \centering
    % \tiny
    \caption{\textbf{Quantitative comparison on the NeRF-DS~\citep{yan2023nerf} dataset with our labeled dynamic specular masks.} We report PSNR, SSIM, and LPIPS (VGG) of previous methods on dynamic specular objects using the dynamic specular objects mask generated by Track Anything~\citep{yang2023track}. The \colorbox{red!25}{best}, the \colorbox{orange!25}{second best}, and \colorbox{yellow!25}{third best} results are denoted by red, orange, yellow. }
    \label{tab:dynamic_spec_tab}
    \resizebox{\textwidth}{!}{%
    \begin{tabular}{lccccccccccccccc}
    \toprule
     & \multicolumn{3}{c}{As} & &\multicolumn{3}{c}{Basin}& &\multicolumn{3}{c}{Bell}& &\multicolumn{3}{c}{Cup} \\ \cmidrule{2-4} \cmidrule{6-8} \cmidrule{10-12} \cmidrule{14-16}
    Method & PSNR↑ & SSIM↑& LPIPS↓& &PSNR↑ & SSIM↑& LPIPS↓& &PSNR↑ & SSIM↑& LPIPS↓& &PSNR↑ & SSIM↑& LPIPS↓  \\ 
    \midrule
    Deformable 3DGS~\citep{yang2023deformable} &\colorbox{orange!25}{24.14}  &\colorbox{yellow!25}{0.7432 } &\colorbox{orange!25}{0.2957} & &17.45  &0.5530  &\colorbox{orange!25}{0.3138} & &\colorbox{orange!25}{19.42}  &\colorbox{orange!25}{0.5516} &\colorbox{orange!25}{0.2940} & &\colorbox{orange!25}{20.10}  &\colorbox{red!25}{0.5446} &\colorbox{red!25}{0.3312}   \\ 
    4DGS~\citep{wu20234d} &22.70  &0.6993  &\colorbox{yellow!25}{0.3517 }& &16.61 &0.4797  &\colorbox{yellow!25}{0.4084} & &14.64  &\colorbox{yellow!25}{0.2596} &\colorbox{yellow!25}{0.4467 }& &18.90  &0.4132 &0.4032   \\
    GaussianShader~\citep{jiang2023gaussianshader} &19.27  &0.5652  &0.5232 & &15.71 &0.4163  &0.5941 & &12.10  &0.1676 &0.6764 & &14.90  &0.3634 &0.6146   \\
    GS-IR~\citep{liang2023gs} &19.32  &0.5857  &0.4782 & &15.21 &0.4009  &0.5644 & &12.09  &0.1757 &0.6722 & &14.80  &0.3445 &0.6046   \\
    NeRF-DS~\citep{yan2023nerf}&\colorbox{yellow!25}{23.67 }&\colorbox{orange!25}{0.7478}  &0.3635 & &\colorbox{orange!25}{17.98} &\colorbox{yellow!25}{0.5537} &0.4211  & &\colorbox{yellow!25}{14.73} &0.2439 &0.5931  & &\colorbox{yellow!25}{19.95} &\colorbox{yellow!25}{0.5079 }&\colorbox{yellow!25}{0.3494}\\
    HyperNeRF~\citep{park2021hypernerf}&17.37  &0.6934  &0.3834 & &\colorbox{red!25}{18.75} &\colorbox{orange!25}{0.5671}  &0.4125 & &13.93  &0.2292 &0.6051 & &15.07  &0.4860 &0.4183   \\
    Ours &\colorbox{red!25}{24.51} &\colorbox{red!25}{0.7534}  &\colorbox{red!25}{0.2896}  & &\colorbox{yellow!25}{17.71 }&\colorbox{red!25}{0.5675} & \colorbox{red!25}{0.3048}&  &\colorbox{red!25}{19.60} &\colorbox{red!25}{0.5680} &\colorbox{red!25}{0.2862}  & &\colorbox{red!25}{20.13} &\colorbox{orange!25}{0.5384} & \colorbox{orange!25}{0.3368}    \\ 
    \midrule
     & \multicolumn{3}{c}{Plate} & &\multicolumn{3}{c}{Press}& &\multicolumn{3}{c}{Sieve}& &\multicolumn{3}{c}{\textbf{Mean}} \\ \cmidrule{2-4} \cmidrule{6-8} \cmidrule{10-12} \cmidrule{14-16}
    Method & PSNR↑ & SSIM↑& LPIPS↓& &PSNR↑ & SSIM↑& LPIPS↓& &PSNR↑ & SSIM↑& LPIPS↓& &PSNR↑ & SSIM↑& LPIPS↓  \\ 
    \midrule
    Deformable 3DGS~\citep{yang2023deformable} &\colorbox{orange!25}{16.12}  &\colorbox{orange!25}{0.5192}  &\colorbox{orange!25}{0.3544} & &19.64 &\colorbox{orange!25}{0.6384} &\colorbox{orange!25}{0.3268} & &\colorbox{red!25}{20.74}  &\colorbox{orange!25}{0.5283} &\colorbox{red!25}{0.3109 }&   &\colorbox{orange!25}{19.66} &\colorbox{orange!25}{0.5826} &\colorbox{orange!25}{0.3181}   \\ 
    4DGS~\citep{wu20234d} &13.93  &0.4095  &0.4229 & &\colorbox{orange!25}{20.17} &0.5434  &\colorbox{yellow!25}{0.4339} & &19.70  &0.4498 &\colorbox{yellow!25}{0.3879} & &18.09  &0.4649 &\colorbox{yellow!25}{0.4078}   \\
    GaussianShader~\citep{jiang2023gaussianshader} &9.87  &0.2992  &0.6812 & &16.84 &0.4408  &0.6093 & &16.19  &0.3241 &0.5862 & &14.98  &0.3681 & 0.6121  \\
    GS-IR~\citep{liang2023gs} &11.09  &0.3254  &0.6270 & &16.43 &0.4083  &0.5776 & &16.42  &0.3339 &0.5749 & &15.05  &0.3678 &0.5856   \\
    NeRF-DS~\citep{yan2023nerf} &14.80 &0.4518  &0.3987 & &\colorbox{yellow!25}{19.77 }&\colorbox{yellow!25}{0.5835}  &0.5035 & &\colorbox{yellow!25}{20.28}  &\colorbox{yellow!25}{0.5173 }&0.4067 & &\colorbox{yellow!25}{18.74}  &\colorbox{yellow!25}{0.5151} &0.4337   \\
    HyperNeRF~\citep{park2021hypernerf}&\colorbox{yellow!25}{16.03 } &\colorbox{yellow!25}{0.4629}  &\colorbox{yellow!25}{0.3775} & &14.10 &0.5365  &0.5023 & &18.39  &\colorbox{red!25}{0.5296} &0.3949 & &16.23  &0.5007 &0.4420   \\
    Ours &\colorbox{red!25}{16.53}  &\colorbox{red!25}{0.5369}  &\colorbox{red!25}{0.3041} & &\colorbox{red!25}{21.70} &\colorbox{red!25}{0.6630}  &\colorbox{red!25}{0.3252} & &\colorbox{orange!25}{20.36}  &0.5089 &\colorbox{orange!25}{0.3190} & &\colorbox{red!25}{20.08}  &\colorbox{red!25}{0.5909} &\colorbox{red!25}{0.3094}   \\ 
    \bottomrule
    \label{table1}
    \end{tabular}
    }
\end{table*}
\begin{figure*}[t]
    \centering
    \includegraphics[width=1\linewidth]{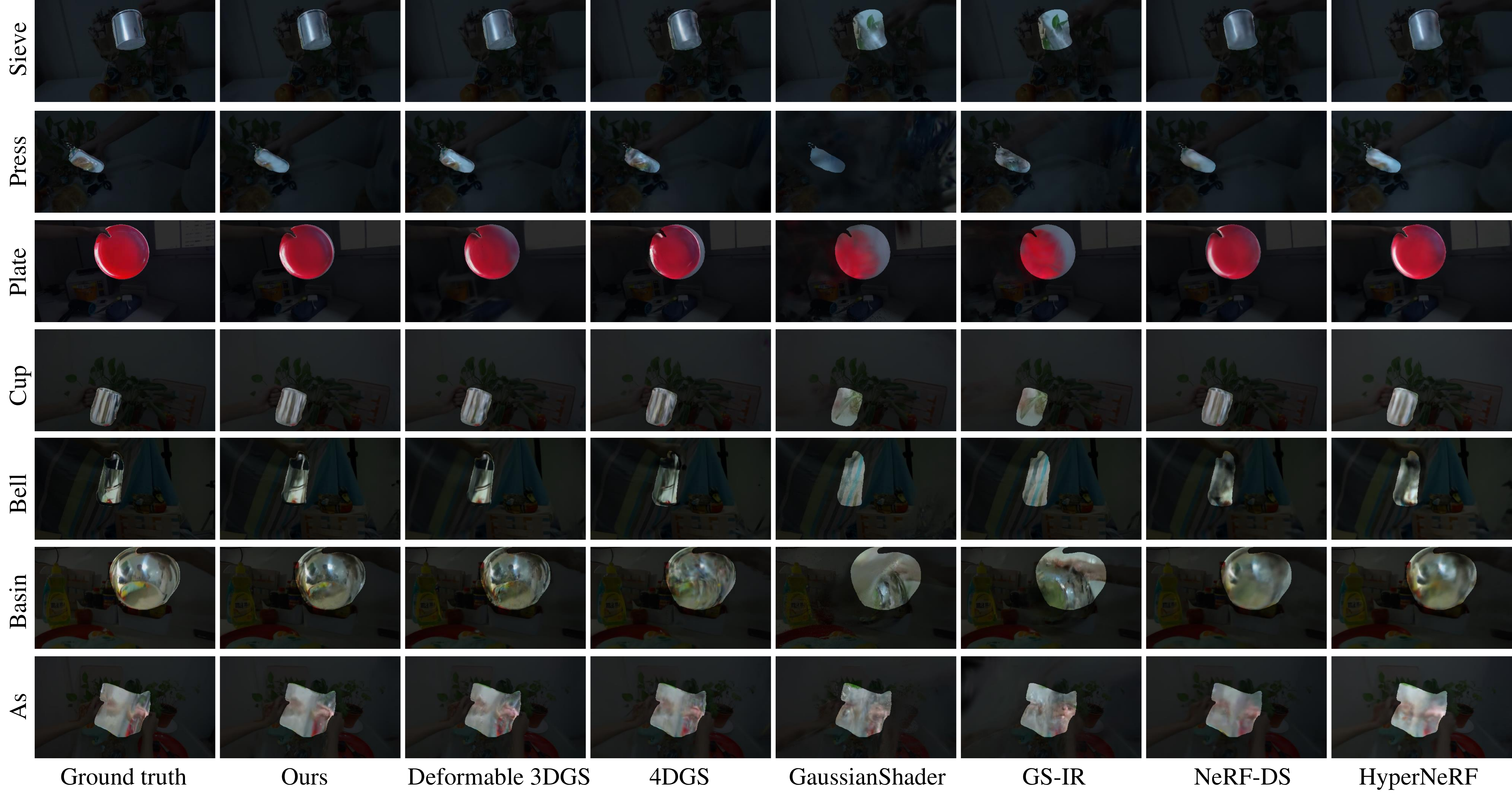}
    \caption{\textbf{Qualitative comparison on NeRF-DS~\citep{yan2023nerf} dataset with labeled dynamic specular masks.}}
    \label{fig:dynamic_spec}
\end{figure*}
Since each scene in the NeRF-DS dataset \citep{yan2023nerf} contains not only dynamic specular objects but also static background objects, we use Track Anything \citep{yang2023track} to obtain masks for the dynamic specular objects. This allows us to evaluate only the dynamic specular objects. As shown in Tab. \ref{tab:dynamic_spec_tab} and Fig. \ref{fig:dynamic_spec}, our method outperforms baselines when evaluating the dynamic specular objects in these monocular sequences.

\subsection{Per-Scene results on the NeRF-DS Dataset}
\begin{figure*}[t]
    \centering
    \includegraphics[width=1\linewidth]{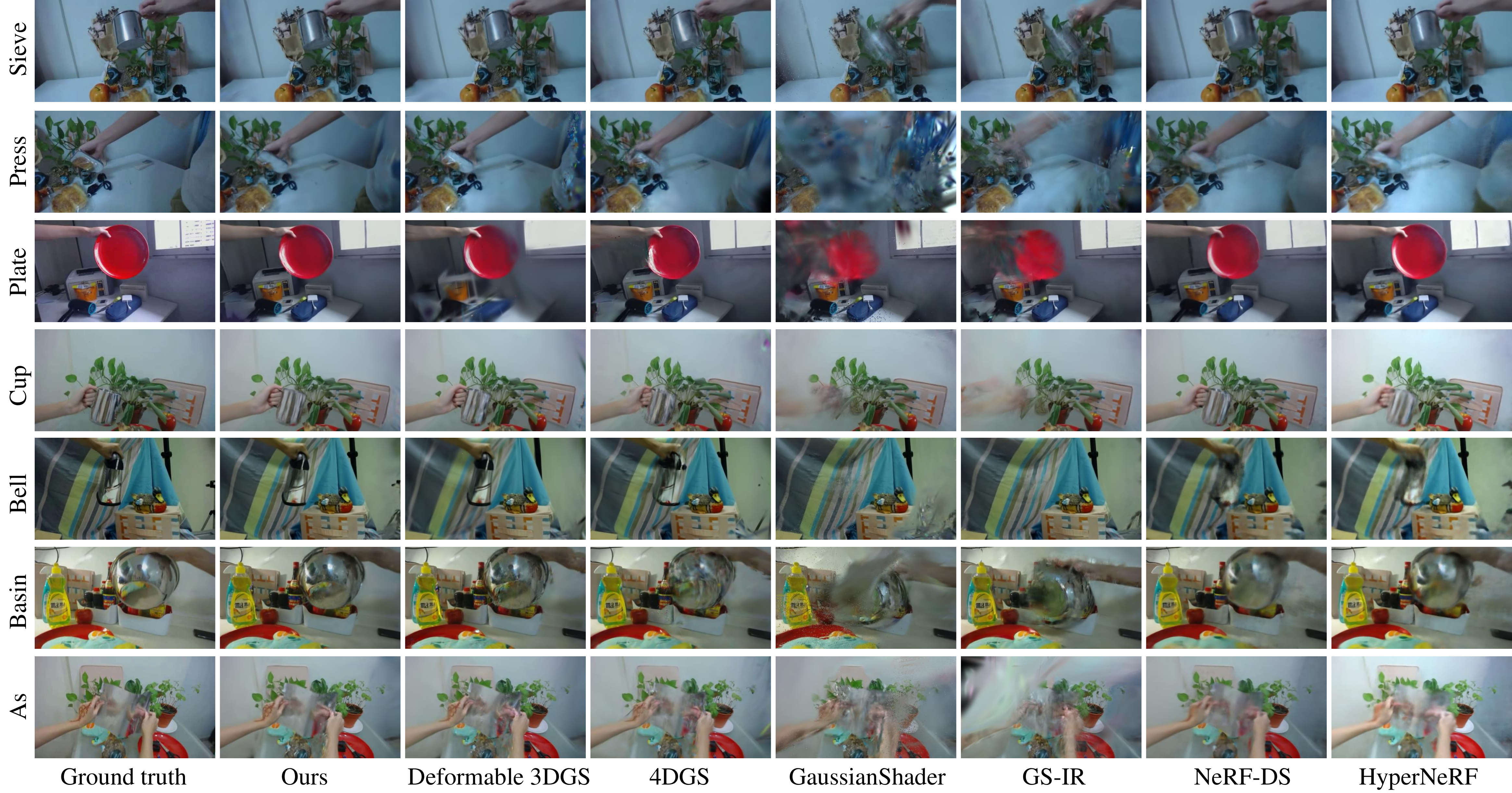}
    \caption{\textbf{Qualitative comparison on the NeRF-DS~\cite{yan2023nerf} dataset.}}
    \label{fig:whole_scene_appendix}
\end{figure*} 
\begin{figure*}[t]
    \centering
    \includegraphics[width=1\linewidth]{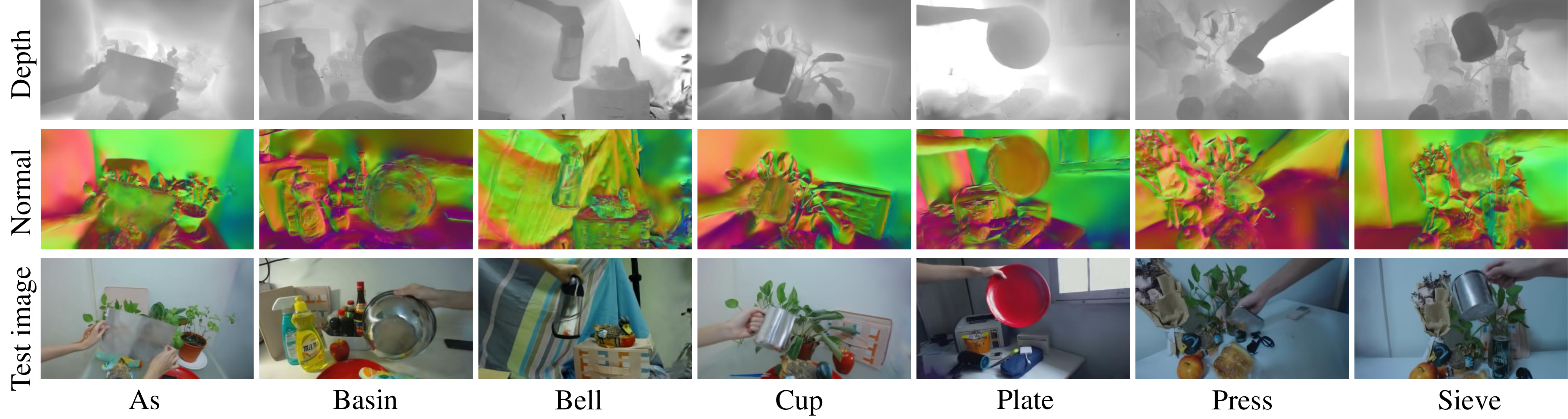}
    \caption{\textbf{Visualized our rendered test images, normal maps, and depth maps.}}
    \label{fig:rgb_normal_depth}
\end{figure*} 
In Fig. \ref{fig:whole_scene_appendix}, we present qualitative results for each scene in the NeRF-DS dataset \citep{yan2023nerf}. The visualizations demonstrate that our method achieves superior rendering quality compared to other approaches. We also provide rendered test images and their corresponding normal maps and depth maps for each scene in the NeRF-DS dataset in Fig. \ref{fig:rgb_normal_depth}.

\subsection{Deformation magnitudes and color decomposition}
\begin{figure*}[t]
    \centering    \includegraphics[width=1.0\linewidth]{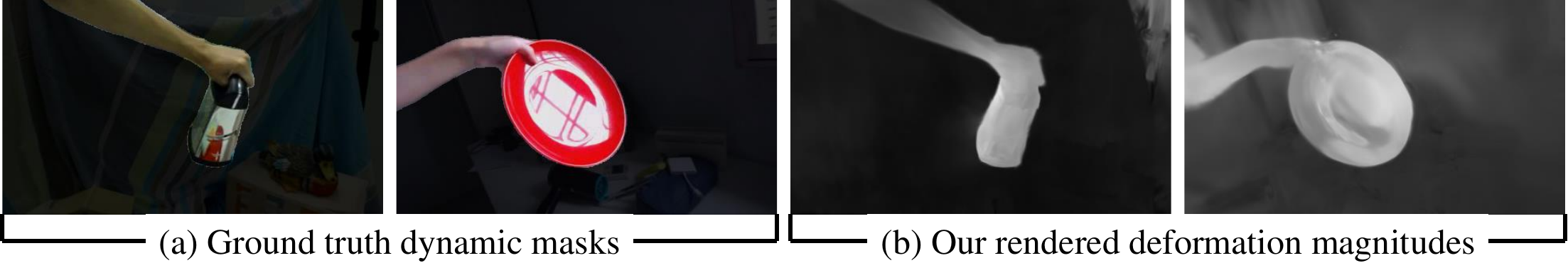}
    \caption{\textbf{Visualized our deformation magnitudes.} (a) The left side shows the ground truth of the dynamic object, while (b) on the right side, we render the magnitude of the output of the position residual by our deformable Gaussian MLP. The brighter areas indicate greater movement of the 3D Gaussians. The figure shows that even without mask supervision, our method can still effectively distinguish which objects are dynamic.}
    \label{fig:dynamic_mask}
\end{figure*}
\begin{figure*}[t]
    \includegraphics[width=1\linewidth]{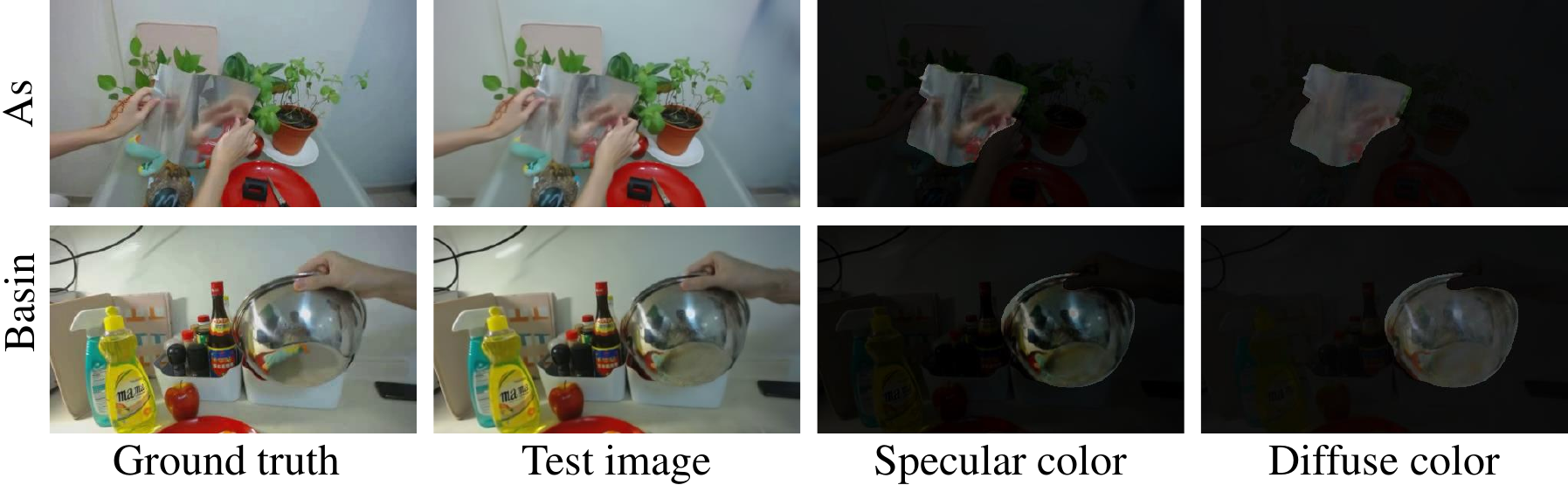}
    \caption{\textbf{Visualized our specular and diffuse color.} Specular regions are emphasized while non-specular areas are dimmed to highlight the results of specular region color decomposition.}
    \label{fig:decompose}
\end{figure*} 
Unlike NeRF-DS \citep{yan2023nerf}, our approach does not require mask supervision to clearly distinguish between static and dynamic objects, as illustrated in Fig.~\ref{fig:dynamic_mask}. Additionally, Fig.~\ref{fig:decompose} illustrates our method's decomposition results. As shown, our approach consistently achieves a realistic separation of specular and diffuse components across different scenes in the NeRF-DS dataset \citep{yan2023nerf}.

\subsection{Training and rendering efficiency} 
\begin{table}[t]
\centering
\small
\caption{\textbf{Training and rendering efficiency on NeRF-DS~\citep{yan2023nerf} dataset}}
\label{tab:fps}
\begin{tabular}{lccc}
    \toprule
     Scene &  Training time (hr) & FPS & Number of Gaussians (k) \\
    \midrule
    as &1.2  &31  &170  \\
    basin &1.3  &25  &218  \\
    bell &1.9  &18  &335  \\
    cup &1.7  &30 &177  \\
    plate &1.1  &27 &187  \\
    press &1.0  &31 &172  \\
    sieve &1.1  &29 &178  \\
    \bottomrule
\end{tabular}
\end{table}
In Tab. \ref{tab:fps}, we present the training time, FPS, and number of Gaussians from our experiments on each scene in the NeRF-DS dataset \citep{yan2023nerf}. The results show that for scenes with fewer than 178k Gaussians, our method achieves real-time rendering greater than or equal to 30 FPS. The experiments are conducted on an NVIDIA RTX 4090 GPU.

\section{Limitation}~\label{sec:limit}
\begin{figure*}[t]
    \centering
    \includegraphics[width=0.8\linewidth]{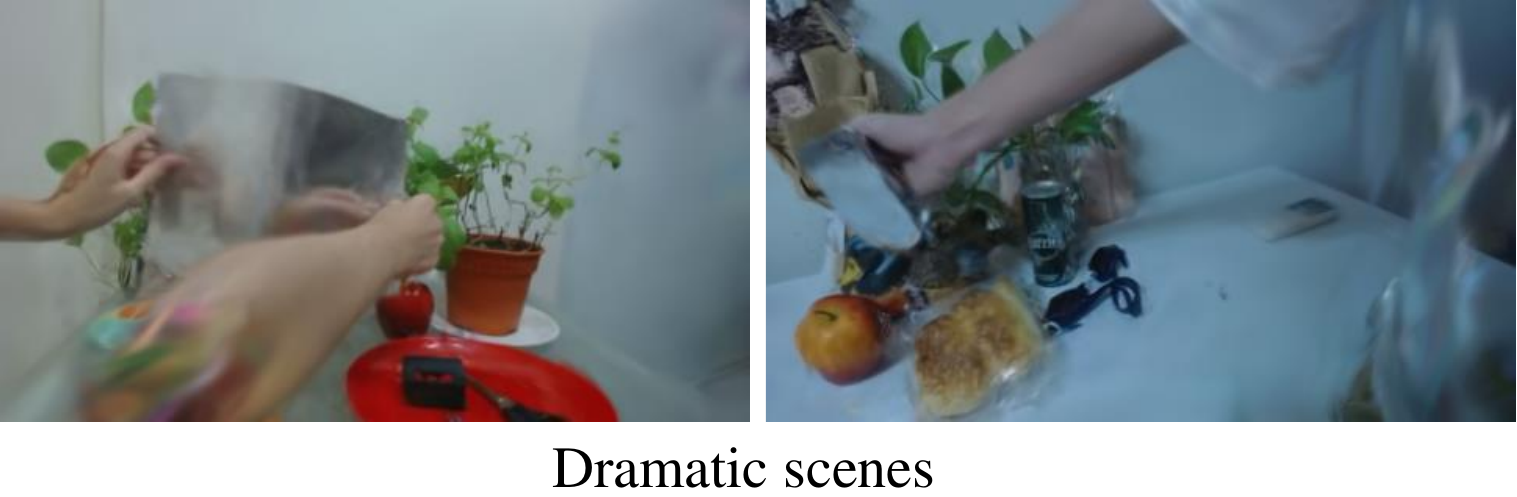}
    \caption{\textbf{Failure cases of modeling dramatic scene changes.} There are dramatic scenes where an arm or body enters or exits the scene, leading to many floaters occurring.}
    \label{fig:limit}
\end{figure*}
In some dramatic scenes, relying solely on the deformable Gaussian MLP and coarse-to-fine training strategy is insufficient, such as when an arm or body enters or exits the scene, leading to many floaters occurring. We provide visual results in Fig. \ref{fig:limit}.

 \fi

\end{document}